%% file: main.tex
\documentclass[10pt,twocolumn,letterpaper]{article}

\usepackage[pagenumbers]{cvpr} % To force page numbers, e.g. for an arXiv version


\definecolor{cvprblue}{rgb}{0.21,0.49,0.74}
\usepackage[pagebackref,breaklinks,colorlinks,allcolors=cvprblue]{hyperref}

\usepackage[ruled,vlined]{algorithm2e}
\usepackage{amsmath}
\usepackage{mathrsfs}
\usepackage{graphicx}
\usepackage{amssymb}
\usepackage{pifont}
\usepackage{tabularx}
\usepackage{bm} 
\usepackage{mathrsfs}
\usepackage{caption}
\usepackage{subcaption}
\usepackage[flushmargin]{footmisc}

\makeatletter
\def\@fnsymbol#1{\ensuremath{}}
\renewcommand{\@makefnmark}{}
\makeatother
\def\paperID{3138} % *** Enter the Paper ID here
\def\confName{CVPR}
\def\confYear{2025}

\title{Erase Diffusion: Empowering Object Removal Through Calibrating \\ Diffusion Pathways}

\author{
    Yi Liu\footnotemark[1]$^{*}$, 
    Hao Zhou\footnotemark[1]$^{*}$, 
    Wenxiang Shang, 
    Ran Lin, 
    Benlei Cui\footnotemark[2]$^{\dagger}$ \\
Alibaba Group \\
{\tt\small \{ly402089,baishu.zh,cuibenlei.cbl,shangwenxiang.swx,linran.lr09\}@taobao.com}
}

\begin{document}

\twocolumn[{
\maketitle
\begin{center}
    \captionsetup{type=figure}
    \includegraphics[width=1.0\textwidth]{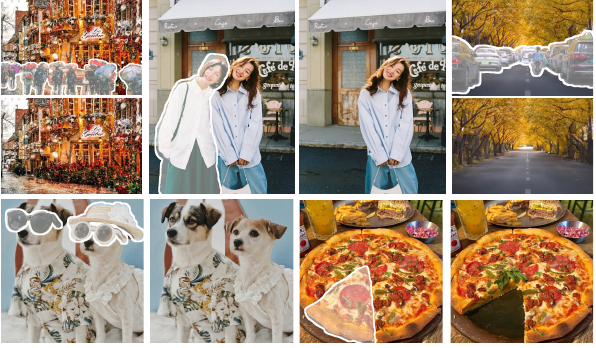}
    \caption{Diverse erase inpainting results produced by our proposed EraDiff, where images before and after removal are presented in pairs, and the areas to be erased in the original images have been marked. The EraDiff can eliminate targets in various complex real-world scenes while ensuring visual coherence in the generated images.}
   \label{fig:teaser}
\end{center}
}]
\maketitle

% 定义脚注内容（必须在\maketitle之后）
\footnotetext[1]{$^{*}$ \scriptsize These authors contributed equally to this work.}
\footnotetext[2]{$^{\dagger}$ \scriptsize Corresponding author.}

\input{sec/0_abstract}

\input{sec/1_introduction}

\input{sec/2_related_work}
\input{sec/3_preliminaries}
\input{sec/3_approach}
\input{sec/4_experiments}

\input{sec/5_conclusion}

% WARNING: do not forget to delete the supplementary pages from your submission 
{
    \small
    \bibliographystyle{ieeenat_fullname}
    \bibliography{main}
}
\input{sec/X_suppl}

\end{document}

%% file: sec/0_abstract.tex
\begin{abstract}
Erase inpainting, or object removal, aims to precisely remove target objects within masked regions while preserving the overall consistency of the surrounding content. Despite diffusion-based methods have made significant strides in the field of image inpainting, challenges remain regarding the emergence of unexpected objects or artifacts. We assert that the inexact diffusion pathways established by existing standard optimization paradigms constrain the efficacy of object removal. To tackle these challenges, we propose a novel Erase Diffusion, termed EraDiff, aimed at unleashing the potential power of standard diffusion in the context of object removal. In contrast to standard diffusion, the  EraDiff adapts both the optimization paradigm and the network to improve the coherence and elimination of the erasure results.
We first introduce a Chain-Rectifying Optimization (CRO) paradigm, a sophisticated diffusion process specifically designed to align with the objectives of erasure. This paradigm establishes innovative diffusion transition pathways that simulate the gradual elimination of objects during optimization, allowing the model to accurately capture the intent of object removal. Furthermore, to mitigate deviations caused by artifacts during the sampling pathways, we develop a simple yet effective Self-Rectifying Attention (SRA) mechanism. The SRA calibrates the sampling pathways by altering self-attention activation, allowing the model to effectively bypass artifacts while further enhancing the coherence of the generated content. With this design, our proposed EraDiff achieves state-of-the-art performance on the OpenImages V5 dataset and demonstrates significant superiority in real-world scenarios.

\end{abstract}

%% file: sec/1_introduction.tex
\section{Introduction}
\label{sec:intro}

% Image inpainting is a crucial computer vision technique designed to reconstruct missing or damaged parts of an image while preserving visual coherence and context \cite{ElHarroussAAA20}. 
% It has diverse applications, such as removing unwanted objects \cite{CriminisiPT03}, restoring damaged photos \cite{LiuRSWTC18}, and editing images \cite{SC-FEGAN}. 
% However, effectively filling in these parts with realistic pixel values remains a significant challenge \cite{YehCLHD16,deepfill}.
Image inpainting is a crucial technique in computer vision, aimed at reconstructing missing or damaged regions while maintaining visual coherence and contextual integrity \cite{3D,aas}. As one specialized form of image inpainting, object removal, also called erase inpainting, has broader applications in various fields such as social media~\cite{appl3,zhou2021embracing}, advertising design~\cite{appl2}, and image processing~\cite{appl1,appl6}. However, in comparison to general image inpainting, an ideal erase inpainting model must address two critical challenges. \textbf{Coherence}: The model should seamlessly inpaint the masked area, ensuring consistency in lighting, content, and other relevant aspects. \textbf{Elimination}: The model should accurately remove objects within the masked area while preventing the generation of extraneous elements or artifacts.

Prior erase inpainting methods primarily relied on non-parametric patch sampling techniques \cite{PatchMatch,LevinZW03,HaysE07} or Generative Adversarial Networks (GANs) \cite{CoModGAN, FCF, LaMa, HiFill, MAT, MI-GAN, SH-GAN, ZITS, ZITS++}. These approaches often resort to filling large masked regions with repetitive patches, resulting in a lack of coherence in the generated images. In recent years, methods based on Latent Diffusion Models (LDMs) \cite{LDMs} have demonstrated superiority in generating more natural images. However, these methods struggle to eliminate target objects. For example, when a user seeks to erase a piece of pizza in Figure~\ref{fig:teaser}, the LDMs may mistakenly generate another piece of pizza instead of the expected clean plate.

We attribute the above issues to the fact that standard diffusion pathways in most existing methods are not suitable for erase tasks. During optimization, most diffusion-based erase inpainting models take the original image with added noise and randomly generated masks as input, aiming to recover the original image in the presence of noise. However, this standard training paradigm only establishes a denoising process that transitions from random noise to clear images, without addressing the specific goal of object removal. Consequently, the model may follow a denoising pathway from noise to an image with objects. Based on this observation, we argue that an ideal erase inpainting model should establish a diffusion pathway directly from objects to backgrounds to ensure the definitive removal of unwanted elements. 
Furthermore, in the early stages of denoising, the reconstructed regions are significantly influenced by the shape of the masks and the level of noise. This impact may lead to deviations in the early latent states, resulting in the emergence of artifacts. The standard self-attention mechanism tends to incorrectly regard the features of these artifacts as important information, which gradually amplifies the deviations during the subsequent denoising process. Ultimately, it leads to unexpected objects in the generated images.

In this paper, we present a novel erase diffusion model, termed \textbf{EraDiff}, specifically designed for object removal, which effectively unleashes the potential erasure  power of standard diffusion. First, we introduce a \textbf{C}hain-\textbf{R}ectifying \textbf{O}ptimization (CRO) paradigm  that supports the establishment of new diffusion pathways from noise to backgrounds. Specifically, we develop a dynamic image synthesis strategy that enables the generation of a variety of dynamic images at different time steps without the need for extra data. These synthesized dynamic images effectively simulate the gradual elimination of objects during the denoising process, yielding intermediate latent states corresponding to different time steps. Next, we introduce a new dedicated optimization objective for erase inpainting. This objective guides the establishment of transitions among multiple intermediate latent states. By applying the CRO paradigm, the model learns to accurately identify the intention to erase and improve content coherence. Furthermore, to address potential deviations caused by artifacts during the early denoising process, we design a \textbf{S}elf-\textbf{R}ectifying \textbf{A}ttention (SRA) mechanism that explicitly guides the model in executing object removal more effectively. By altering self-attention activation, the SRA  enhances background features while rectifying its incorrect reliance on artifacts. This replacement of the standard attention mechanism results in final generated images that appear more realistic and devoid of unexpected objects.  To the best of our knowledge, our proposed method achieves state-of-the-art performance in both the public OpenImages V5~\cite{OpenImages} dataset and large-scale real-world scenarios.

%% file: sec/2_related_work.tex
\section{Related Work}
\label{sec:related_work}

%-------------------------------------------------------------------------
% \subsection{Image Inpainting}

\noindent \textbf{Image Inpainting.}
Image inpainting, which involves the reconstruction of large-scale missing regions, has garnered substantial attention within the field.
Traditional image inpainting methods employ heuristic patch-based propagation algorithms to borrow texture and structure from neighboring regions of the corrupted areas \cite{PatchMatch,LevinZW03,HaysE07}. However, these methods often struggle to effectively restore areas with complex structures or semantics \cite{MAGIC}. 
To tackle this issue, numerous studies have harnessed the power of deep neural networks \cite{dl_1, dl_2, zhou2019visual} and proposed various enhancements to GANs \cite{dl_4} aimed at improving the coherence of inpainted regions. 
Subsequent research has extensively examined different facets, including semantic context and texture \cite{dl_sematic_1,dl_sematic_2,dl_sematic_3}, edges and contours \cite{ZITS,ZITS++}, as well as the design of hand-engineered architectures \cite{dl_engineer_1,dl_engineer_2}. Notably, Suvorov et al. \cite{LaMa} introduced FFCs into residual modules, providing a remedy for enhancing the perceptual quality of GAN-generated outputs. 
More recently, promising efforts have been made in applying LDMs to inpainting tasks, showcasing their potential to achieve impressive results \cite{RePaint, SmartBrush,inp2,mo_ldm1,mo_ldm2,mo_ldm3}.
However, general image inpainting models may generate unexpected objects in the masked areas, making these LDM-based methods unsuitable for effective object removal.

% In this study, we address this challenge by employing a semi-supervised training approach combined with a masked self-attention mechanism during the denoising process, significantly enhancing the reliability and adaptability of LDMs for this task.

%-------------------------------------------------------------------------
% \subsection{Erase Inpainting}

\noindent \textbf{Erase Inpainting.}
Erase inpainting, one specialized form of image inpainting, focuses on removing unwanted content from images.  The typical setup for erase inpainting is to provide an image along with an object mask as input, ultimately generating a clean background that excludes the specified object. Many studies have concentrated on improvements in model architecture \cite{HiFill,MAT,DreamInpainter,Magicremover} and loss functions \cite{MI-GAN,LaMa}. While these methods effectively improve the coherence of masked regions, they still struggle with object removal. Recent works \cite{InstInpaint,EmuEdit,ZONE,bld,SmartEdit} have attempted to leverage textual prompts to locate erase targets within images. However, relying on textual cues to accurately identify areas for erasure has proven to be unstable. Models fail to follow textual instructions, which can lead to unfavorable results. As a result, these approaches face challenges in large-scale practical applications. Through in-depth investigation, we find that the struggle to effectively erase objects stems from the inexact construction of the erase diffusion chain. Our proposed method attempts to directly establish diffusion pathways from the noise to the background. This approach enables effective object removal without relying on additional textual information and produces one clean and natural background.

%% file: sec/3_preliminaries.tex
\section{Preliminaries}
\label{sec:preliminaries}

\noindent \textbf{Latent Diffusion Models.} 
In this paper, we utilize latent diffusion models (LDMs) as our erase inpainting model. LDMs are probabilistic models designed to learn a data distribution \(p(\bm{x})\) by gradually denoising a normally distributed variable, which corresponds to learning the reverse sampling process of a fixed Markov Chain of length \(T\). Specifically, the forward diffusion process progressively adds noise to the initial latent state \(\bm{x}_0\),
\begin{equation}
\label{eq:1}
    q(\bm{x}_t|\bm{x}_0) = \mathcal{N}(\bm{x}_t;\sqrt{\bar{\alpha_t}} \bm{x}_0,(1-\bar{\alpha_t})\bm{I}),
\end{equation}
where \(1-\bar{\alpha}_t\) is the variance schedule at step \(t\) and represents the level of noise, with \(\bar{\alpha}_t = \prod_{s=1}^t\alpha_s\).

According to the DDIM algorithm, the reverse sampling process can be defined as

\begin{equation}
\label{eq:2}
\begin{aligned}
    \bm{x}_{prev} = & \sqrt{\bar{\alpha}_{prev}} \left( \frac{\bm{x}_{t} - \sqrt{1 - \bar{\alpha}_{t}} \epsilon_{\theta}^{(t)}(\bm{x}_{t})}{\sqrt{\bar{\alpha}_{t}}} \right) \\
    & + \sqrt{1 - \bar{\alpha}_{prev}-\sigma_{t}^2} \epsilon_{\theta}^{(t)}(\bm{x}_{t}) + \sigma_{t}\epsilon_{t},
\end{aligned}
\end{equation}
where \(\epsilon_{t}\sim \mathcal{N}(\bm{0}, \bm{I})\) is standard Gaussian noise. Here, \(\sigma_{t}\) are hyper-parameters, and the reverse sampling process can be considered deterministic as \(\sigma_{t} \to 0\). The term \(\epsilon_{\theta}^{(t)}(\bm{x}_{t})\) is trained to predict the noise added to \(\bm{x}_t\) by minimizing the following optimization objective
\begin{equation}
\label{eq:3}
\min_{\theta} \mathbb{E}_{\epsilon \sim  \mathcal{N}(\bm{0}, \bm{I})} \left\| \epsilon - \epsilon_{\theta}^{(t)}(\bm{x}_{t}) \right\|_2^{2}.
\end{equation}

\noindent \textbf{Structure of Erasing Ipainting.} 
Most advanced erasing inpainting methods are based on diffusion models. In this paper, we focus on the SD2-Inpaint model\cite{LDMs}, which mainly consists of a Variational AutoEncoder (VAE) \cite{vae} and a U-Net network \cite{unet}. The VAE facilitates the transformation of input images into latent space. For Erasing Ipainting, the inputs typically consist of a noisy image \(\bm{x}_t^{ori} \in \mathbb{R}^{H \times W \times 3}\), a random binary mask \( \textbf{M} \in \{0, 1\}^{H \times W} \), and the corresponding masked image \( \bm{x}'= \bm{x}_0^{ori} \odot\textbf{M}\). The U-Net network predicts the associated noise at timestamp \(t\) based on the Equation~\ref{eq:3}. This methodology enables the model to establish a diffusion chain between the original image and Gaussian noise, allowing for seamlessly repainting masked regions. 

Additionally, the self-attention mechanism within U-Net networks plays a significant role in the erasing process. This mechanism aggregates information from the entire image, thereby controlling the generation of features in the masked regions during the reverse sampling process. 
Given one latent feature map \(\bm{z} \in \mathbb{R}^{h \times w \times c}\), where \(h\), \(w\) and \(c\) are the height, width, and channel dimensions of \(z\) respectively, the self-attention process can be represented as follows
\begin{equation}
\label{eq:4}
    \mathbf{Q}, \mathbf{K}, \mathbf{V} = \ell_Q(\bm{z}), \ell_K(\bm{z}), \ell_V(\bm{z}),
\end{equation}
\begin{equation}
\label{eq:5}
    \mathrm{SA}(\mathbf{Q},\mathbf{K},\mathbf{V}) = \mathrm{Softmax}\left(\frac{\mathbf{Q} \mathbf{K}^\top}{\sqrt{d}}\right) \mathbf{V},
\end{equation}
where \(\ell_Q,\ell_K,\ell_V\) are learnable linear layers, and \( d \) denotes the scaling factor.

%% file: sec/3_approach.tex
\section{Method}
\label{sec:method}
\begin{figure}[t]
    \centering
    \includegraphics[width=0.475\textwidth]{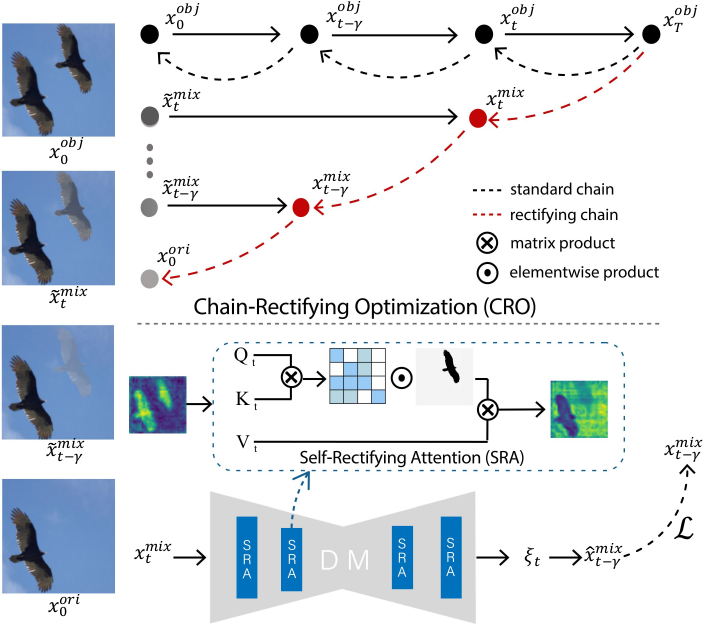}
    \caption{The overview of our proposed Erase Diffusion, termed EraDiff. \textbf{Left}: Dynamic image synthesis. Each image is initially transformed using techniques like matting, scaling, and copy-pasting. A mix-up strategy then synthesizes a series of dynamic images \{$\bm{\tilde{x}}_{t}^{mix}$\} that simulate the gradual fading of the object. \textbf{Top}: Chain-Rectifying Optimization (CRO). The standard sampling pathway is prone to generating artifacts (black dashed lines). In contrast, we establish a new sampling path for erasing (red dashed lines) that better aligns the reverse sampling trajectory with a clear background. \textbf{Bottom}: Self-Rectifying Attention (SRA). The standard self-attention mechanism may inadvertently amplify artifacts, diverging from the expected diffusion pathway. By modifying the attention activation, we guide the model to bypass artifact regions, enhancing its focus on the background and ensuring a more accurate erase sampling path.}
    \label{fig:framework}
\end{figure}

\noindent \textbf{Overview.} 
The overall architecture of our proposed Erase Diffusion, termed EraDiff, is shown in Figure~\ref{fig:framework}, consisting of a \textbf{C}hain-\textbf{R}ectifying \textbf{O}ptimization paradigm (see Section \ref{sec:method:optimization}) and a \textbf{S}elf-\textbf{R}ectifying \textbf{A}ttention mechanism (see Section \ref{sec:method:attention}). 
This CRO paradigm introduces the design of dynamic latent states along with a dedicated optimization objective, thereby establishing novel diffusion chains that better align with the erasure objectives. Meanwhile, the SRA mechanism effectively guides the sampling process to prevent state deviations caused by artifacts.

\subsection{Chain-Rectifying Optimization}
\label{sec:method:optimization}
In the optimization process, existing erase inpainting methods typically employ randomly generated or object-based masks, predicting the noise in the masked regions using Equation~\ref{eq:3}. These masked regions may cover objects, and the model's optimization objective is to reconstruct these objects rather than eliminate them.  As discussed above, such standard diffusion chains may inadvertently encourage the model to generate unexpected artifacts.
An effective solution is to dedicate diffusion paths for the erase inpainting task between objects and backgrounds. By simulating the gradual fading of objects, this approach can suppress their emergence during the sampling process. Even if the sampling process does not start with standard Gaussian noise (a common enhancement trick), it can still help eliminate leaked objects within the masked region. Thus, the main challenge of this approach is to develop new latent states within the erase diffusion chains and to optimize the model to manipulate transitions among these latent states.

\noindent \textbf{Dynamic Latent States.} 
To design dedicated diffusion chains for erasure between objects and backgrounds, it is essential to obtain corresponding image pairs. However, existing public datasets often lack these paired samples, highlighting the need for an innovative data synthesis strategy.

Let \(\bm{x}_0^{ori}\) denote the original image, and a trained matting model is employed to segment the primary object within it. We then apply various image transformation techniques, including rotation and scaling, to modify the segmented objects. These transformed objects are randomly pasted onto the background region, resulting in a new synthesized image, \(\bm{x}_0^{obj}\). This method allows us to construct a training dataset of object-background image pairs at a relatively low cost. Furthermore, to effectively simulate the gradual fading of the objects, we input dynamic images \(\bm{\tilde{x}}_{t}^{mix}\) for each time step \(t\) during the training process, 
\begin{equation}
\label{eq:6}
    \bm{\tilde{x}}_{t}^{mix} = (1-\lambda_t) \bm{x}_{0}^{ori}+ \lambda_t \bm{x}_{0}^{obj},
\end{equation}
where the decreasing sequence \(\lambda_{:T} \in [0, 1]^{T}\) controls the mix-up level of object-background image pairs. In the case of \(t = 0\), \(\bm{\tilde{x}}_{0}^{mix}\) corresponds to the original image \( \bm{x}_{0}^{ori}\). These dynamic mix-up images assist the model in understanding the smooth transition from objects to backgrounds. Ultimately, based on Equation~\ref{eq:1}, we can derive a set of new latent states \(\bm{x}_{t}^{mix}\) for the self-rectifying diffusion chains,
\begin{equation}
\label{eq:7}
    \bm{x}_{t}^{mix} =\sqrt{\bar{\alpha}_t}\bm{\tilde{x}}_{t}^{mix} + \sqrt{1-\bar{\alpha_t}} \epsilon, 
\end{equation}
where \(\epsilon \sim \mathcal{N}(\bm{0}, \bm{I})\).

\noindent \textbf{Optimization Objective.} 
The traditional diffusion process starts from a fixed \(\bm{x}_0^{ori}\), and each intermediate latent state \(\bm{x}_t^{ori}\) can be obtained directly using Equation~\ref{eq:1}. However, in the new self-rectifying diffusion chains, each latent state \(\bm{x}_t^{mix}\) is derived from the dynamic image \( \bm{\tilde{x}}_t^{mix}\), making it infeasible to apply the standard optimization objective in Equation~\ref{eq:3}. To mitigate this challenge, we propose a novel optimization objective that enhances the alignment between the model's predicted distribution and the true distribution, as illustrated in Algorithm~\ref{algo:cre}.

\begin{algorithm}
\caption{Chain-Rectifying Optimization}
\label{algo:cre}
\SetNlSty{textbf}{}{} % Optional: style for line numbers
 % Adjust size of line numbers if needed
\Repeat{converged}{
    $\bm{x}_{0}^{ori} \sim q(x)$\;
    $t \sim \text{Uniform}(\{1, \ldots, T\})$\;
    $\gamma \sim \text{Uniform}(\{1,\ldots,\gamma_{m}\})$\;
    $\epsilon \sim \mathcal{N}(0, I)$\;
    $\bm{x}_{0}^{obj} \longleftarrow \bm{x}_{0}^{ori}$ image transformation\;
    $\bm{x}_{t}^{mix}, \bm{x}_{t-\gamma}^{mix} \longleftarrow \bm{x}_{0}^{ori},\bm{x}_{0}^{obj} \text{ based on Equ}~\ref{eq:6}-\ref{eq:7}$\;
    $\bm{\hat{x}}_{t-\gamma}^{mix} \longleftarrow \bm{x}_{t}^{mix},\epsilon_{\theta}^{(t)} \text{ based on Equ}~\ref{eq:8}$\;
    Take gradient descent step on 
    \vspace{-1em}
    \begin{equation*}
    \nabla_\theta \left\| \bm{x}_{t-\gamma}^{mix} - \bm{\hat{x}}_{t-\gamma}^{mix}\right\|^2 \text{ based on Equ}~\ref{eq:9}
    \end{equation*}
    \vspace{-1.7em}
    }
\end{algorithm}

Specifically, the new loss function aims to minimize the distance between the model-predicted latent states \(\bm{\hat{x}}_t^{mix}\) and the true states \(\bm{x}_t^{mix}\). It allows the model to gradually alter its parameters to adapt to the new distribution shifts. Given one latent state \(\bm{x}_t^{mix}\), we can derive the model-predicted latent state \(\bm{\hat{x}}_{t-\gamma}^{mix}\) at the previous \(\gamma\) time step,
\begin{equation}
\label{eq:8}
\begin{aligned}
    p_\theta(\bm{\hat{x}}_{t-\gamma}^{mix}|\bm{x}_{t}^{mix} ) = & \sqrt{\bar{\alpha}_{t-\gamma}} \left( \frac{\bm{x}_{t}^{mix} - \sqrt{1 - \bar{\alpha}_{t}} \epsilon_{\theta}^{(t)}(\bm{x}_{t}^{mix})}{\sqrt{\bar{\alpha}_{t}}} \right) \\
    & + \sqrt{1 - \bar{\alpha}_{t-\gamma}} \epsilon_{\theta}^{(t)}(\bm{x}_{t}^{mix}),
\end{aligned}
\end{equation}
where \(\gamma \in (0,\gamma_m)\). Given the original image \(\bm{x}_0^{ori}\) and the synthesized image \(\bm{x}_0^{obj}\), the corresponding true state \(\bm{x}_{t-\gamma}^{mix}\) can be obtained using Equation~\ref{eq:6} and \ref{eq:7}. Finally, we define the new optimization objective as follows
\begin{equation}
\label{eq:9}
\min_{\theta} \mathbb{E}_{\gamma \sim  \text{Uniform}(1, \gamma_{m}),t} \left\| \bm{x}_{t-\gamma}^{mix} - p_\theta(\bm{\hat{x}}_{t-\gamma}^{mix}|\bm{x}_{t}^{mix} ) \right\|_2^{2}.
\end{equation}

\subsection{Self-Rectifying Attention}
\label{sec:method:attention}
We have established a self-rectifying diffusion pathway to guide the model toward object removal. However, information leakage from the mask's shape can still lead to artifacts during the early stages of denoising, resulting in latent state shifts. The self-attention mechanism tends to give the masked region stronger attention to itself rather than to the background. This phenomenon can continuously amplify artifacts along the reverse sampling path, ultimately leading to a deviation from the object removal direction. An intuitive solution is to alter the current attention layers for path calibration to mitigate the above risk of generating unexpected objects due to deviations from specific states.

Based on our observations, we believe that the generation of the foreground region should rely more on backgrounds rather than focusing on itself. Additionally, the background region remains visible and thus should not be affected by the content of the foregrounds. By altering the self-attention activation, it is possible to ignore the negative effects of artifacts while emphasizing the background, thereby further enhancing the coherence and elimination of the generated content. Thus, we propose a simple yet effective Self-Rectifying Attention mechanism to replace the standard self-attention mechanism. Specifically, we first downsample and flatten the image mask  \(\textbf{M}\) to obtain the corresponding mask vector \(m \in \{0, 1\}^{wh}\). We then design an extended mask \(m' \in \{-inf, 1\}^{wh \times wh}\) as follows
\begin{equation}
\label{eq:10}
m'_{i,j} = 
\begin{cases} 
1, & m_{i}=0~\text{or}~m_{j}=0  \\ 
-inf, & \text{else}
\end{cases}
\end{equation}
The extended mask is subsequently applied directly to the corresponding attention activations, effectively suppressing objects within the masked regions while enhancing background features. This is represented mathematically as
\begin{equation}
\label{eq:csa}
    \mathrm{SRA}(\mathbf{Q},\mathbf{K},\mathbf{V}) = \mathrm{Softmax}\left(\frac{\mathbf{Q} \mathbf{K}^\top}{\sqrt{d}} \cdot m'\right) \mathbf{V}.
\end{equation}
This mechanism enhances the model's capacity to perceive background information without any additional computational cost, while also effectively diminishing the interference of artifacts on the final generated results. Consequently, this improvement ensures that the sampling process can calibrate itself to the target direction of object removal.

%% file: sec/4_experiments.tex
\section{Experiments}
\label{sec:experiments}

\subsection{Experimental Setup}
\label{sec:exp:details}
\noindent \textbf{Benchmark Datasets.} 
We conducted a thorough evaluation of our proposed pipeline utilizing the publicly available OpenImages V5 segmentation dataset~\cite{OpenImages}. Each instance within this dataset comprises the original image, a corresponding segmentation mask, segmentation bounding boxes, and associated class labels, thereby enabling a rigorous comparative analysis against several baseline models. For our experiments, we randomly selected a subset of 10,000 samples from the OpenImages V5 test set.

\noindent \textbf{Comparison Baselines.} 
We have chosen several cutting-edge image inpainting methods as our baselines for comparison. These include two mask-guided approaches: SD2-Inpaint~\cite{LDMs} and LaMa~\cite{LaMa}, alongside two text-guided techniques: Inst-Inpaint~\cite{InstInpaint} and PowerPaint~\cite{powerpaint}. 
% To further validate the effectiveness of our proposed method, we have also integrated SD2.1~\cite{LDMs} into our baselines.

\noindent \textbf{Evaluation Metrics.} 
In accordance with LaMa~\cite{LaMa}, we employ two primary evaluation metrics: Fréchet Inception Distance (FID)~\cite{fid} and Learned Perceptual Image Patch Similarity (LPIPS)~\cite{lpips}. These metrics are adept at assessing the overall visual coherence of the inpainted images. To provide a more nuanced evaluation of the quality of content generated within the masked regions, we introduce the Local FID metric~\cite{local_fid}, which allows for a detailed assessment of local visual fidelity. Furthermore, we augment our evaluation framework by incorporating analyses derived from GPT-4o~\cite{gpt} along with expert human annotations, thereby examining the effectiveness of these erasure-targeted models in eliminating objects and artifacts.

\noindent \textbf{Training Details.} 
Our proposed pipeline builds upon the foundations established by SD2-Inpaint and is implemented using PyTorch along with the Diffusers library. During the training phase, we employed the Adam~\cite{Adam} optimizer with a learning rate set to \(3 \times 10^{-6}\) on the OpenImages V5 training set. To simplify parameters, the increasing sequence \(\lambda_{:T}\) follows the same schedule as the sequence \(1-\bar{\alpha}_{:T}\). For loss computation, we ensured that the time intervals between two timestamps did not exceed \(\gamma_m = 100\). All experiments were conducted on NVIDIA A100 GPUs.
%%%%%%
\begin{figure*}[t]
    \centering
    \includegraphics[width=\textwidth]{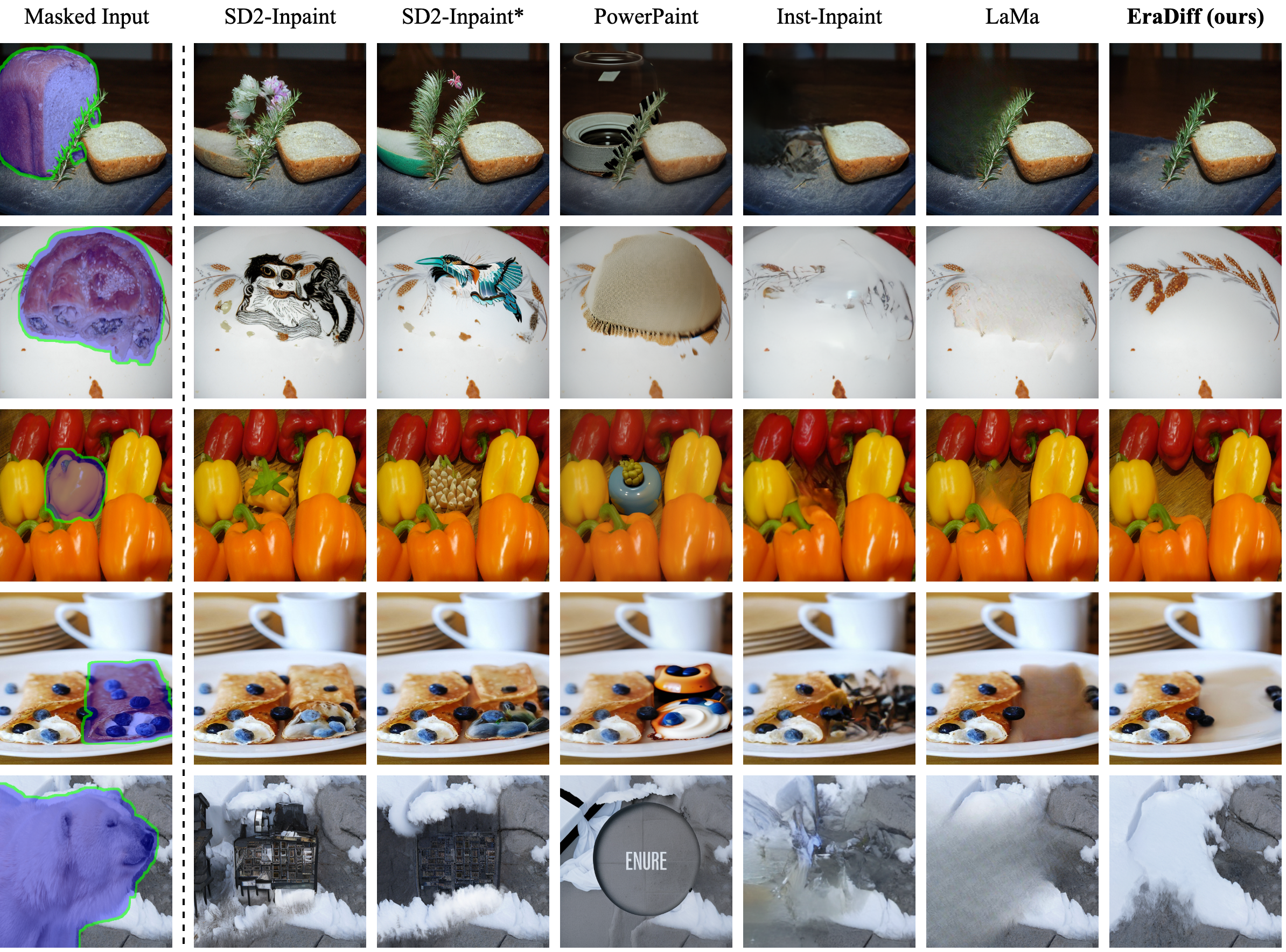}
    \caption{Qualitative results of OpenImages V5 dataset compared among SD2-Inpaint~\cite{LDMs}, SD2-Inpaint with prompt guidance~\cite{LDMs}, PowerPaint~\cite{powerpaint}, Inst-Inpaint~\cite{InstInpaint}, LaMa~\cite{LaMa}, and our approach.}
    \label{fig:vis_coherence}
\vspace{-0.6em}
\end{figure*}
%%%%%

\subsection{Qualitative and Quantitative Comparisons}
\label{sec:exp:coh}
%%%%%
\begin{table}[t]
    \centering
    \begin{tabular}{p{2.3cm} p{1.3cm} p{1.3cm} p{1.7cm}}
        \toprule
        Method & FID$ \downarrow $ & LPIPS$ \downarrow $ & Local FID$ \downarrow $\\
        \midrule
        SD2-Inpaint                     & \textbf{3.805}    & 0.301             & 8.852             \\
        % SD2-Inpaint$^{*\dagger}$        & \underline{4.019} & 0.308
        SD2-Inpaint$^{*\dagger}$        & \underline{4.019} & 0.308 
        & \underline{7.194} \\
        PowerPaint                      & 6.027             & 0.289             & 10.021            \\
        Inst-Inpaint                    & 11.423            & 0.410             & 43.472            \\
        LaMa                            & 7.533             & \underline{0.219} & 6.091             \\
        \textbf{EraDiff (ours)}                   & 6.540             & \textbf{0.192}    & \textbf{3.799}    \\
        \bottomrule
    \end{tabular}
    \raggedright % 左对齐
    % \footnotesize{ 
    % $^{\dagger}$ In this experiment, we transformed the class labels corresponding to the masked regions into prompts that serve as guiding information for the denoising process of SD2-Inpaint.
    % }
\caption{Quantitative assessment of various erase inpainting models on the OpenImages V5 dataset. Optimal results are highlighted in bold, with runner-up performance underlined.}
\label{tab:main}
\vspace{-0.6em}
\end{table}
%%%%%%%
The quantitative evaluation results of our proposed method, in comparison to several established models, are summarized in Table~\ref{tab:main}. 
Additionally, we present the experimental findings when the prompt is used as supplementary guidance for SD2-Inpaint (abbreviated as SD2-Inpaint*, consistent throughout the paper). 
As indicated in the table, our model is in the mid-range of all baseline models in terms of the FID score, yet it significantly surpasses the others in the Local FID metric. This observation highlighting its remarkable ability to generate visually coherent results specifically within the designated erased regions, all the while maintaining a commendable level of visual coherence throughout the entire image. Additionally, our method records the highest performance on the LPIPS metric, suggesting that the images produced exhibit enhanced visual fidelity following the removal process. 
% It is crucial to note that both the FID and LPIPS metrics predominantly assess the visual coherence of the final generated images and do not provide a direct indication of the efficacy of object or artifact elimination. This limitation is exemplified in Figure~\ref{fig:vis_coherence}, which illustrates that while the SD2-Inpaint model achieves the best FID score, it often fails to adequately erase objects present in the masked areas. Furthermore, this figure clearly demonstrating our model's robustness in addressing challenges such as the presence of objects within the image that closely resemble the target erasure region, as well as extensive areas that necessitate removal.
Figure~\ref{fig:vis_coherence} illustrates relevant visualization results, clearly demonstrating the model's robustness in addressing challenges such as the presence of objects within the image that closely resemble the target erasure region, as well as extensive areas that necessitate removal. 
% Notably, the figure highlights that the GAN-based LaMa model tends to produce outputs that are more visually blurred than those generated by all other SD-based models, including our proposed approach.
% It is crucial to note that both the FID and LPIPS metrics predominantly assess the visual coherence of the final generated images and do not provide a direct indication of the efficacy of object or artifact elimination. This limitation is exemplified in Figure~\ref{fig:vis_coherence}, which illustrates that while the SD2-Inpaint model achieves the best FID score, it often fails to adequately erase objects present in the masked areas. Similarly, the LaMa model demonstrates comparable shortcomings in this regard. Furthermore, as depicted in this figure, the GAN-based LaMa model tends to produce outputs that are more visually blurred than those generated by all other SD-based models, including our proposed approach.
%%%%%%%
% \subsection{Elimination Analysis}
%%%%%
%%%%
\begin{table}[t]
\vspace{-0.2em}
    \centering
    \begin{tabular}{p{2.2cm} p{1.3cm} p{1.7cm} p{1.4cm}}
        \toprule
        % Method & $\mathcal{S}\uparrow$ & $\mathcal{E}$ & $\mathcal{I}\downarrow$\\
        Method & Superior & Comparable & Inferior\\
        \midrule
        SD2-Inpaint              & 1.03\%       & 18.07\%             & 80.90\%         \\
        SD2-Inpaint$^{*}$        & 2.20\%       & 24.79\%             & 73.01\%         \\
        LaMa                     & 13.17\%      & 35.29\%             & 51.54\%         \\
        \bottomrule
    \end{tabular}
    % \raggedright % 左对齐
    % \footnotesize{ 
    % $^{\dagger}$ In this experiment, we transformed the class labels corresponding to the masked regions into prompts that serve as guiding information for the denoising process of SD2-Inpaint.
    % }
\caption{Quantitative results of OpenImages V5 dataset among SD2-Inpaint, SD2-Inpaint$^*$, LaMa, and EraDiff. This table delineates a comparative analysis of the elimination performance results obtained by these methodologies relative to ours, highlighting whether their outcomes are superior, comparable, or inferior to those achieved by our approach.}
\label{tab:gsb}
\end{table}
%%%%%%%%%%
\begin{figure}[t]
    \centering
    \includegraphics[width=0.475\textwidth]{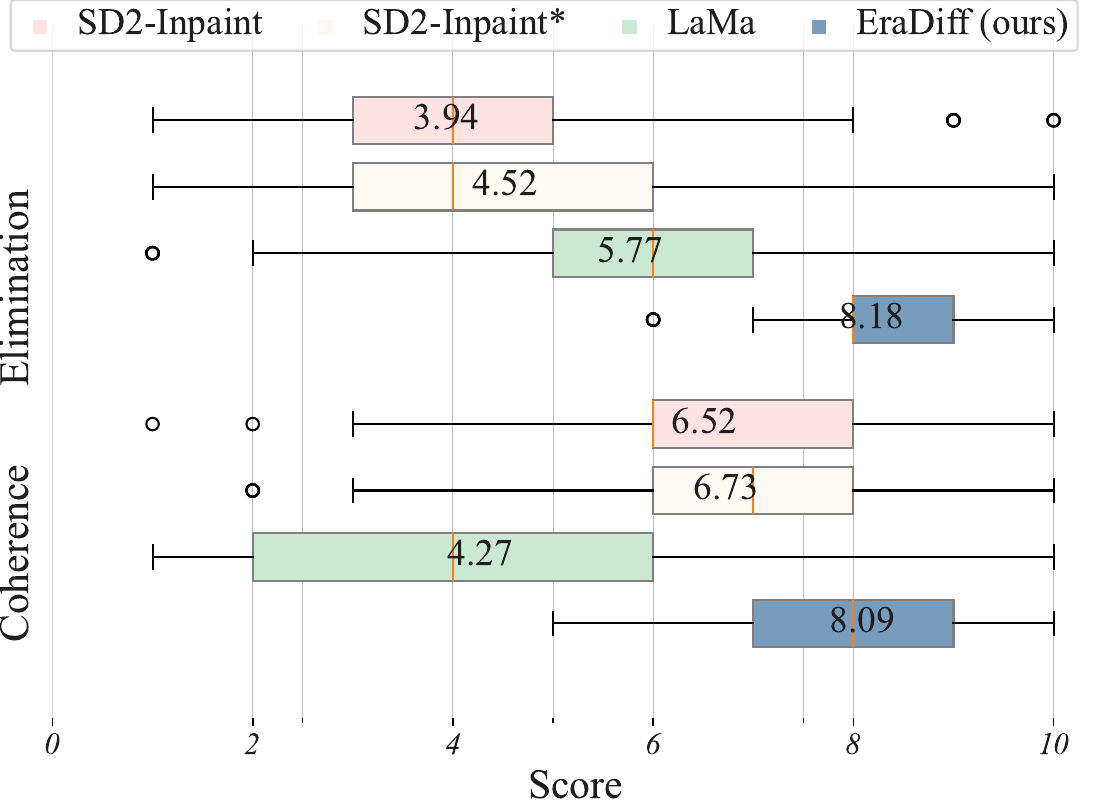}
    \caption{Results from the user study. EraDiff demonstrates enhanced performance, as indicated by its higher mean scores in both elimination and coherence evaluations.}
    \label{fig:plot}
\vspace{-1.em}
\end{figure}
%%%%%%%%%%

It is essential to highlight that both the FID and LPIPS metrics primarily evaluate the visual coherence and aesthetic quality of the final generated images. However, these metrics do not provide a direct assessment of the effectiveness regarding object or artifact elimination in the denoising process. This limitation is clearly illustrated in Figure~\ref{fig:vis_coherence}, which demonstrates that despite the SD2-Inpaint model achieving the highest FID score among all evaluated models, it frequently fails to adequately eliminate unwanted objects or artifacts from the designated masked areas. 
% It is crucial to note that both the FID and LPIPS metrics predominantly assess the visual coherence of the final generated images and do not provide a direct indication of the efficacy of object or artifact elimination. This limitation is exemplified in Figure~\ref{fig:vis_coherence}, which illustrates that while the SD2-Inpaint model achieves the best FID score, it often fails to adequately erase objects present in the masked areas. 

To evaluate the effectiveness of the baseline models and our proposed method in eliminating objects and artifacts, we systematically selected the top-performing models: SD2-Inpaint, SD2-Inpaint*, and LaMa, as indicated in Table~\ref{tab:main}. A comparative analysis was performed through pairwise comparisons against our method, utilizing GPT-4o to generate objective evaluations focused on identifying superior results among the competing models. The outcomes of this analysis are described in Table~\ref{tab:gsb}.
% We constructed a triplet for evaluation, consisting of images generated by each selected model, images produced by our proposed method, and queries derived from the class labels associated with the erased regions. These triplets were subsequently presented to the GPT-4o model, enabling it to assess and identify the superior results among the competing models, as depicted in Figure 2. 
Furthermore, 
% to address potential biases inherent in large models when interpreting complex images and to garner more robust experimental insights, 
we undertook a user study that involved 20 experts that was tasked with appraising the effectiveness of the erasure and the visual aesthetics of the generated results, using a scoring system ranging from 1 to 10; where a score of 1 reflects poor performance and a score of 10 indicates exceptional quality. The outcomes of this assessment are illustrated in Figure~\ref{fig:plot}. 
The results from both experimental methodologies distinctly demonstrate that our proposed approach significantly outperforms the competitors in effectively eliminating target objects. This conclusion is further corroborated by the finding that, despite the lower FID score recorded for SD2-Inpaint, its efficacy in object removal does not surpass that of our innovative model.

Furthermore, we find that EraDiff shows superior performance across a diverse array of in-the-wild scenarios, as illustrated in Figure~\ref{fig:wild}. A more detailed analysis of this will be provided in the appendix. 

%%%%%%%%%%
\begin{figure}[t]
    \centering
    \includegraphics[width=0.475\textwidth]{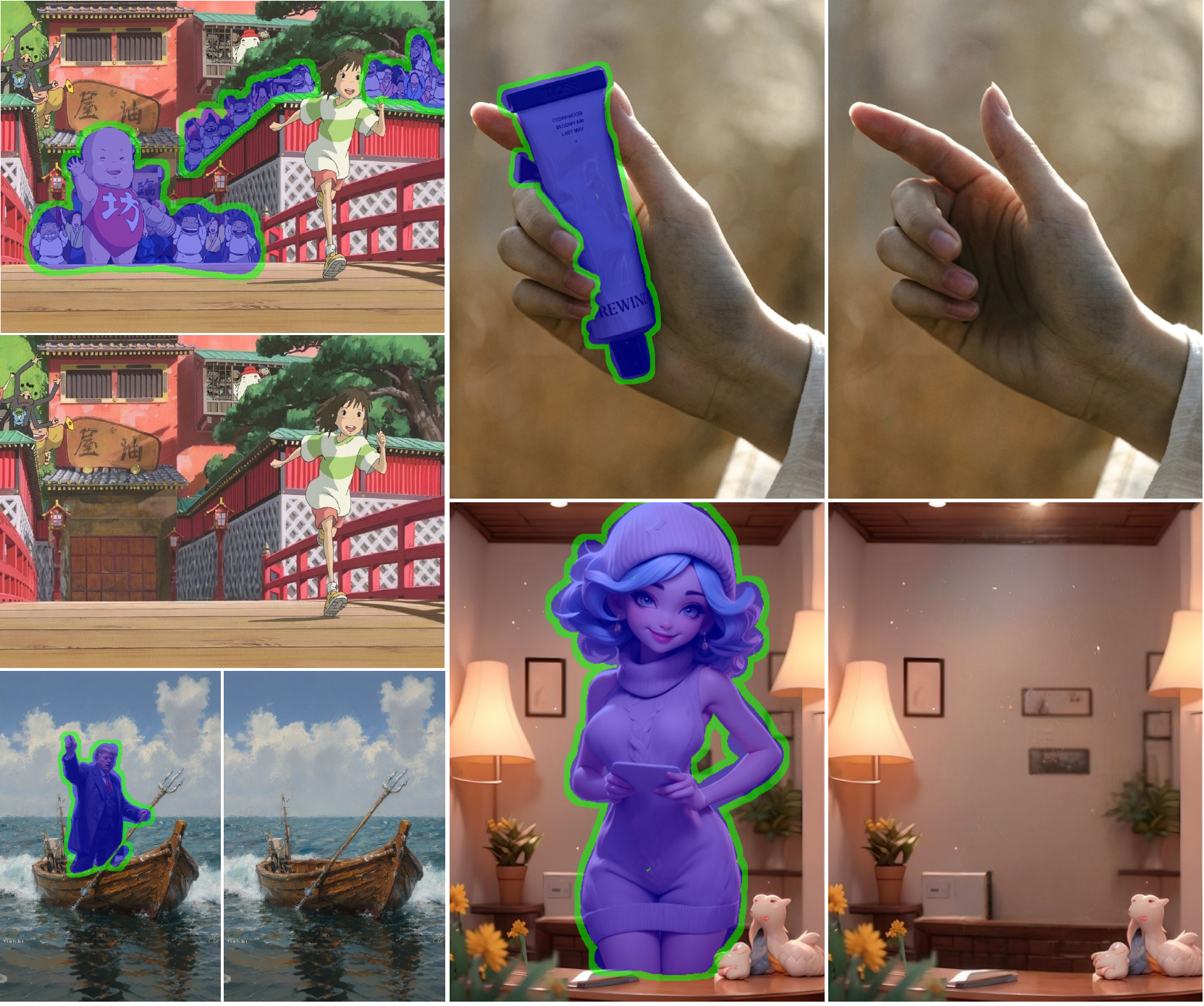}
    \caption{Visualization of EraDiff's performance across a diverse array of in-the-wild scenarios: animated imagery, e-commerce content, oil paintings, and glasses-free 3D visuals.}
    \label{fig:wild}
\vspace{-1.em}
\end{figure}
%%%%%%%%%%
% \label{sec:exp:eli}

\subsection{Ablation Study}
\label{sec:exp:abl}
\begin{figure}[t]
    \centering
    \includegraphics[width=0.475\textwidth]{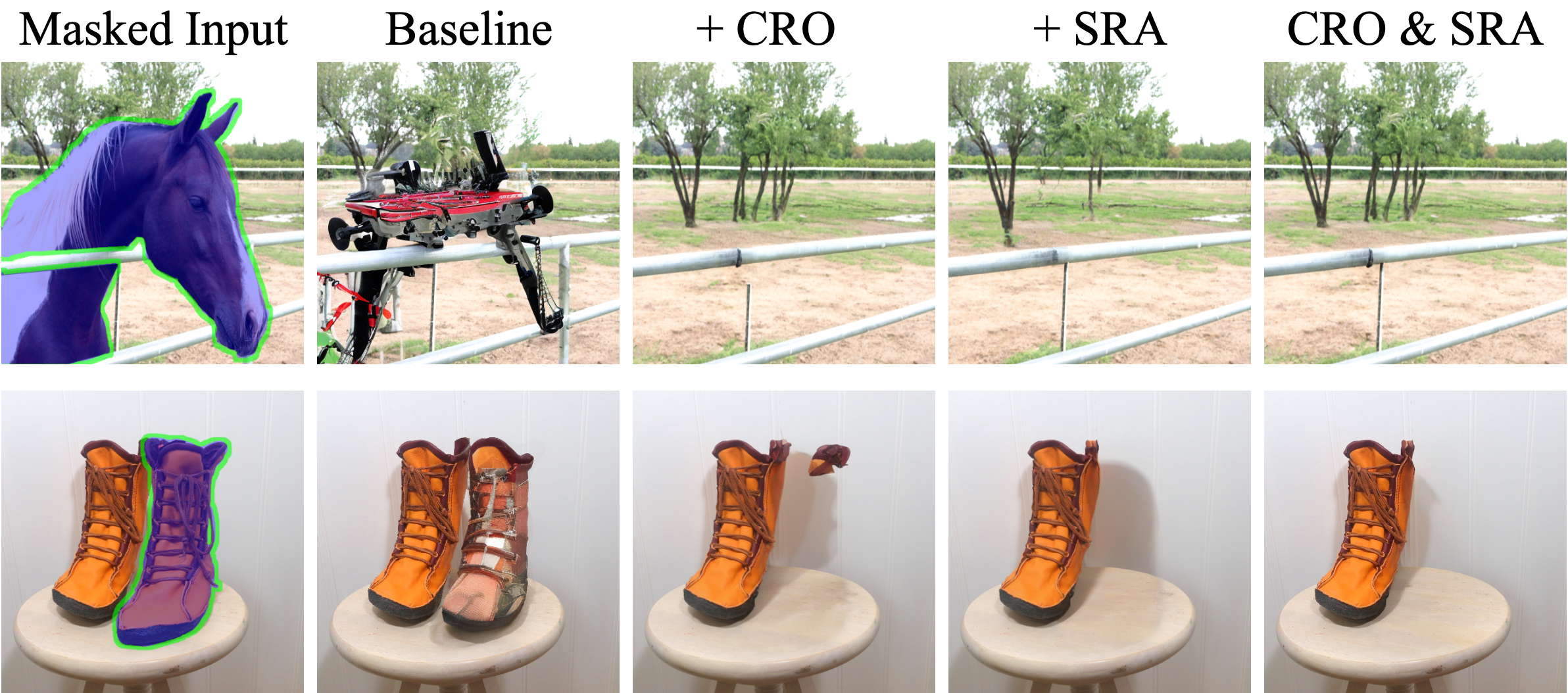}
    \caption{Visual examples for the ablation study comparing baseline, baseline with CRO, baseline with SRA, and baseline with both CRO and SRA, displayed left to right.}
    \label{fig:vis_ablation_main}
\vspace{-0.2em}
\end{figure}

To evaluate the impact of calibrating the sampling pathway on the performance of erase inpainting, we conducted four extensive experiments: each focusing on the removal of CRO, SRA, and the simultaneous exclusion of both CRO and SRA, all based on the EraDiff model. Furthermore, we assessed the model's performance without the incorporation of the mix-up strategy during training, wherein the value of \( \lambda_t \) was held constant. For our evaluation, we employed the GPT-4o to quantify the success rate of object elimination. Results are detailed in Table~\ref{tab:abl}. 
The results indicate that the removal of either the CRO or SRA components, as well as their simultaneous exclusion, leads to a substantial deterioration in both the visual coherence of the erased images and the efficacy of object elimination. Particularly noteworthy is the fact that the omission of CRO results in a more pronounced decline in overall performance. 
Additionally, the absence of the mix-up strategy produced considerable fluctuations in training loss and impeded model convergence. The instability observed can be attributed to the limited noise present at earlier time steps, which complicates the model's ability to accurately predict the entirety of the masked region as noise. 
Figure~\ref{fig:vis_ablation_main} presents the visual outcomes of these ablation experiments, clearly illustrating the effectiveness of the calibration sampling pathway in enhancing the erasure task. In this regard, SRA serves a corrective function, effectively addressing specific anomalous denoising artifacts. 
% Additionally, as outlined in Section~\ref{sec:method:optimization}, we contrasted the outcomes of forgoing the mixup technique in favor of a straightforward pasting method for image blending. 
%%%%%%%%
\begin{table}[t]
% %%%%%%
    \centering
    \begin{tabular}{p{2.7cm} p{1.9cm} p{1.7cm}}
        \toprule
        Method & Local FID$ \downarrow $ & GPT score$ \uparrow $\\
        \midrule
        EraDiff                                 & 3.799                 & 83.43\%     \\
        \ -$w/o\ \text{CRO}$                    & 5.713                 & 72.96\%     \\
        \ -$w/o\ \text{SRA}$                    & 4.950                 & 78.54\%     \\
        \ -$w/o\ \text{CRO} \cup \text{SRA}$      & 8.852                 & 27.80\%     \\
        % Without Mix-up                              & $\mathrm{NaN}$        & $\mathrm{NaN}$       \\
        \ -$w/o\ \text{mix-up}$                             & $\mathrm{NaN}$        & $\mathrm{NaN}$       \\
        \bottomrule
    \end{tabular}
\caption{Results of the ablation study highlight the individual and combined effects of CRO and SRA methodologies.}
\label{tab:abl}
\end{table}
% %%%%%%
% %%%%%%

\subsection{How the CRO and SRA work?}
\label{sec:exp:ana}
To further elucidate the roles of the CRO and SRA methodologies in the denoising process of erase inpainting, we established two types of extended experimental setups. 

\begin{figure}[t]
    \centering
    \includegraphics[width=0.475\textwidth]{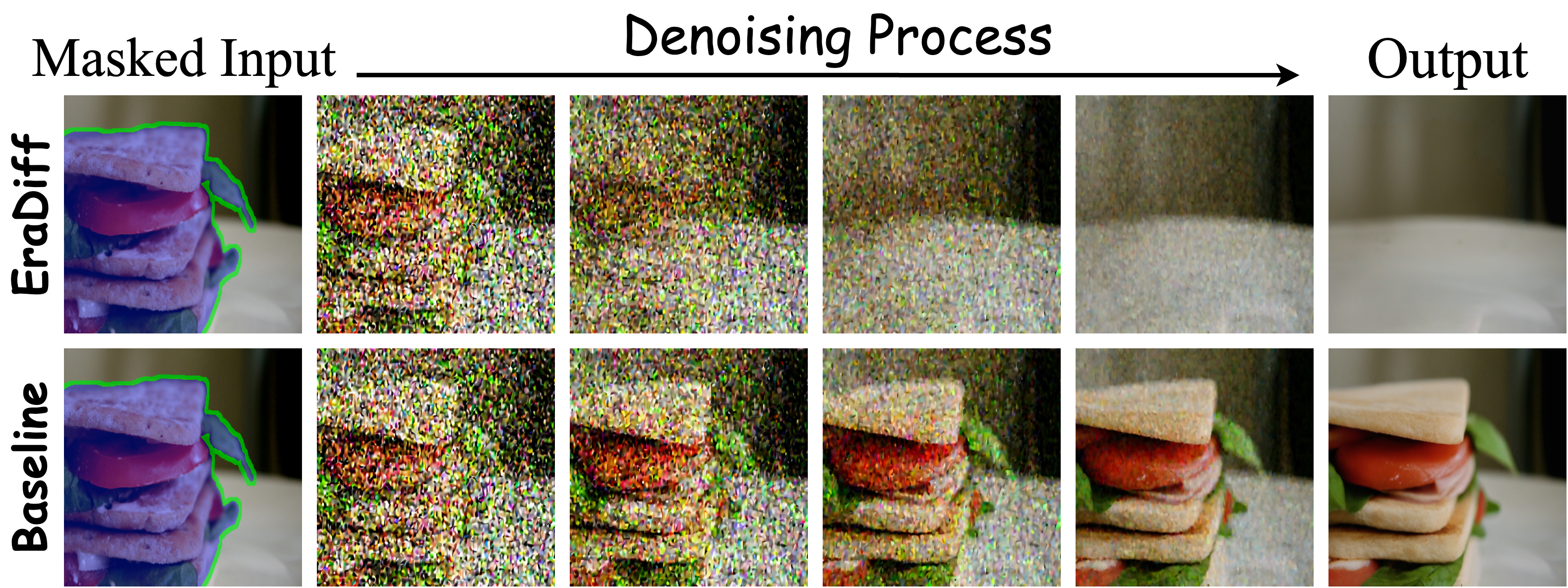}
    \caption{Comparison of EraDiff's and baseline model's denoising process with strength set to 0.6.}
    \label{fig:vis_cro}
\vspace{-0.2em}
\end{figure}

To investigate the mechanism underlying CRO, we established an extremely challenging condition by setting the denoising strength to 0.6. This configuration allows for significant leakage of original image information, including the target objects intended for removal. This leakage can lead to considerable disruptions, resulting in artifacts within the erased regions and potentially even the restoration of the erasure-targeted objects. As illustrated in Figure~\ref{fig:vis_cro}, we observed the denoising processes of the EraDiff alongside the baseline model (\textit{i.e.}, SD2-Inpaint) under this condition. Notably, as the denoising process progressed, the EraDiff model effectively concealed the object artifacts in the erased region, whereas the baseline model highlighted it. These experimental findings provide further validation of our earlier discussion in Section~\ref{sec:method}, asserting that EraDiff accomplishes the erasure task by traversing a sample pathway that closely resembles \( x_t^{mix} \). 

\begin{figure}[t]
    \centering
    \includegraphics[width=0.475\textwidth]{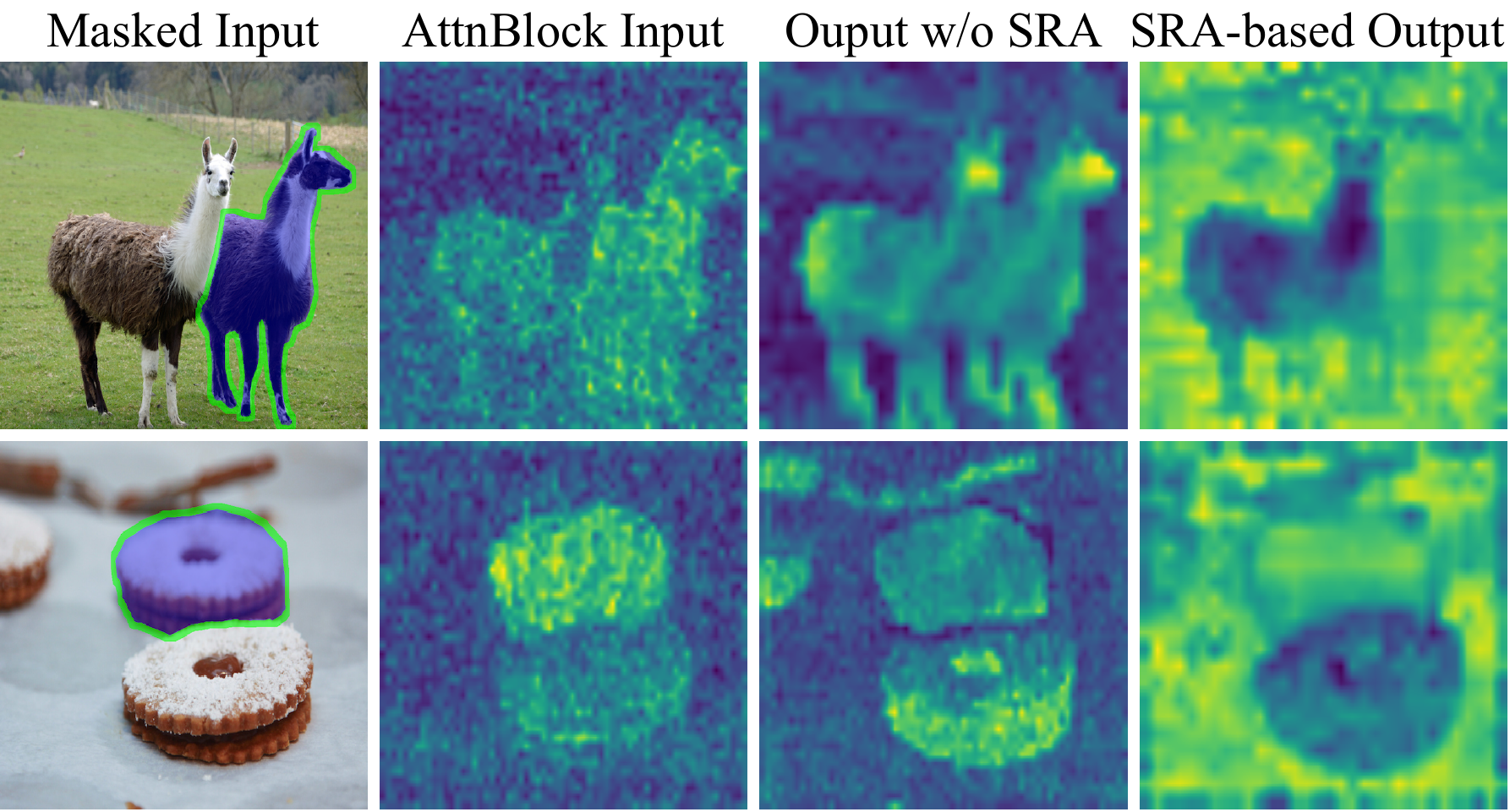}
    \caption{Visualization of heatmaps representing attention block outputs in the presence and absence of the SRA mechanism.}
    \label{fig:vis_sra}
\vspace{-0.2em}
\end{figure}

To reveal the mechanism of the SRA approach, we examined a misleading scenario where multiple similar objects exist within an image, necessitating the removal of one specific object. The heatmaps generated before and after the attention block for both the EraDiff and the baseline model are presented in Figure~\ref{fig:vis_sra}. The results indicate that the baseline model tends to focus on similar objects, while our proposed method shifts its attention to the background. This strategic focus allows the model to extract critical information from the background rather than from the foreground, thereby facilitating the elimination of the target object while maintaining coherence between the erased region and the surrounding background.
%%%%%%
%%%%%%
%%%%%%
%%%%%%

%%%%%%
% \begin{figure}[t]
%     \centering
%     \includegraphics[width=0.47\textwidth]{images/vis_strength.pdf}
%     \caption{Visual comparison with baseline methods.}
%     \label{fig:vis_strength}
% \end{figure}
%%%%%%
% \begin{figure}
%      \centering
%      \begin{subfigure}[b]{0.47\textwidth}
%          \raggedleft
%          \includegraphics[width=\textwidth]{images/vis_ablation.pdf}
%          \caption{$y=x$}
%          \label{fig:y equals x}
%      \end{subfigure}
%      \hfill
%      \begin{subfigure}[b]{0.475\textwidth}
%          \centering
%          \includegraphics[width=\textwidth]{images/vis_strength.pdf}
%          \caption{$y=3\sin x$}
%          \label{fig:three sin x}
%      \end{subfigure}
%      % \hfill
%      % \begin{subfigure}[b]{0.3\textwidth}
%      %     \centering
%      %     \includegraphics[width=\textwidth]{images/vis_strength.pdf}
%      %     \caption{$y=5/x$}
%      %     \label{fig:five over x}
%      % \end{subfigure}
%     \caption{Three simple graphs}
%     \label{fig:three graphs}
% \end{figure}

%%%%

%% file: sec/5_conclusion.tex
\section{Conclusion}
\label{sec:conclusion}

In this paper, we present the EraDiff, which enhances object elimination while maintaining visual coherence in erase inpainting. By introducing a CRO paradigm, EraDiff establishes innovative diffusion pathways that facilitate a gradual removal of objects, allowing the model to better understand the erasure intent. Moreover, the SRA mechanism effectively reduces artifacts during the sampling process.  With inclusive experiments, we demonstrate that these advancements significantly improve performance in challenging scenarios, positioning EraDiff as a valuable contribution to the field of erase inpainting.

%% file: sec/X_suppl.tex
\clearpage
\setcounter{page}{1}
\maketitlesupplementary

% \section{Rationale}
% \label{sec:rationale}
% % 
% Having the supplementary compiled together with the main paper means that:
% % 
% \begin{itemize}
% \item The supplementary can back-reference sections of the main paper, for example, we can refer to \cref{sec:intro};
% \item The main paper can forward reference sub-sections within the supplementary explicitly (e.g. referring to a particular experiment); 
% \item When submitted to arXiv, the supplementary will already included at the end of the paper.
% \end{itemize}
% % 
% To split the supplementary pages from the main paper, you can use \href{https://support.apple.com/en-ca/guide/preview/prvw11793/mac#:~:text=Delete%20a%20page%20from%20a,or%20choose%20Edit%20%3E%20Delete).}{Preview (on macOS)}, \href{https://www.adobe.com/acrobat/how-to/delete-pages-from-pdf.html#:~:text=Choose%20%E2%80%9CTools%E2%80%9D%20%3E%20%E2%80%9COrganize,or%20pages%20from%20the%20file.}{Adobe Acrobat} (on all OSs), as well as \href{https://superuser.com/questions/517986/is-it-possible-to-delete-some-pages-of-a-pdf-document}{command line tools}.

\section{Derivation of Chain-Rectifying Algorithms}
\label{suppl:math}

The Chain-Rectifying Algorithms presented in this study significantly enhance the visual coherence and elimination of artifacts in the erase inpainting task. A critical aspect of this enhancement is the domain transform from the noise \(\epsilon\) predicted by standard diffusion techniques to predicted noise \(\epsilon_\theta\) generated by our method. In the subsequent sections, we will thoroughly elucidate this difference and its implications for the inpainting process.

Based on Equation 6 and Equation 7 in the main text of the paper, we can obtain
\begin{equation}
\label{eqn:supp1:1}
\begin{aligned}
    \bm{x}_{t - 1}^{mix} &= \sqrt{\bar{\alpha}_{t-1}} \left( \bar{\alpha}_{t-1} \bm{x}_0^{ori} + (1-\bar{\alpha}_{t-1}) \bm{x}_0^{obj} \right) \\
    &\quad + \sqrt{1 - \bar{\alpha}_{t-1}} \epsilon.
\end{aligned}
\end{equation}
Additionally, by transforming Equation 7, we can derive
\begin{equation}
\label{eqn:supp1:2}
    \tilde{\bm{x}}_{t}^{mix} = \frac{\bm{x}_{t}^{mix} - \sqrt{1 - \bar{\alpha}_{t}} \epsilon}{\sqrt{\bar{\alpha}_{t}}}.
\end{equation}
Substituting Equation~\ref{eqn:supp1:2} into Equation 7 in the main text and replacing \( \lambda_t \) with \( 1 - \bar{\alpha}_t \) according to the experimental setup, we can obtain
\begin{equation}
\label{eqn:supp1:3}
    \frac{\bm{x}_{t}^{mix} - \sqrt{1 - \bar{\alpha}_{t}} \epsilon}{\sqrt{\bar{\alpha}_{t}}} = \bar{\alpha}_{t} \bm{x}_0^{ori} + (1 - \bar{\alpha}_{t}) \bm{x}_0^{obj}.
\end{equation}
By combining Equation~\ref{eqn:supp1:1}, Equation~\ref{eqn:supp1:2}, and Equation~\ref{eqn:supp1:3}, we can further infer that

\begin{equation}
\label{eqn:supp1:4}
\begin{aligned}
    \bm{x}_{t - 1}^{mix} &= \frac{1}{\alpha_t\sqrt{\alpha_t}} \bm{x}_{t}^{mix} + \frac{\sqrt{\bar{\alpha}_{t-1}} (\alpha_{t}-1)}{\alpha_t} \bm{x}_0^{obj} \\
    &+ (\sqrt{1 - \bar{\alpha}_{t-1}} - \frac{\sqrt{1-\bar{\alpha}_t}}{\alpha_t\sqrt{\alpha_t}}) \epsilon.
\end{aligned}
\end{equation}

According to the DDIM inversion~\cite{ddim}, we can also obtain
\begin{equation}
\label{eqn:supp1:5}
    \bm{x}_{t - 1}^{mix} = \sqrt{\bar{\alpha}_{t-1}} \frac{\bm{x}_{t}^{mix} - \sqrt{1-\bar{\alpha}_t} \epsilon_\theta}{\sqrt{\bar{\alpha}_t}} + \sqrt{1-\bar{\alpha}_{t-1}} \epsilon_\theta,
\end{equation}
where $\epsilon_\theta$ is the prediction of our erase diffusion model.

Finally, by combining Equation~\ref{eqn:supp1:4} and Equation~\ref{eqn:supp1:5}, we can obtain 
\begin{equation}
\label{eqn:supp1:6}
    \text{G}\epsilon_\theta = \text{A}\bm{x}_{t}^{mix} + \text{B}\bm{x}_{0}^{obj} + \text{C}\epsilon,
\end{equation}
where
\[
\text{G} = \sqrt{\alpha_t - \bar{\alpha}_{t}} - \sqrt{1 - \bar{\alpha}_t},
\]
\[
\text{A} = \frac{1}{ \alpha_t} - 1,
\]
\[
\text{B} = \frac{\sqrt{\bar{\alpha}_{t}} (\alpha_{t}-1)}{\alpha_t},
\]
\[
\text{C} = \sqrt{\alpha_t - \bar{\alpha}_{t}} - \frac{\sqrt{1-\bar{\alpha}_t}}{\alpha_t}.
\]
From the equation~\ref{eqn:supp1:6}, we can find that our network adjusts the standard predictions $\epsilon$ by leveraging the object information $\bm{x}_{0}^{obj}$ and the latent state $\bm{x}_{t}^{mix}$.

\section{Comparison with Standard Diffusion}
\label{suppl:comp}
\begin{figure}[t]
    \centering
    \includegraphics[width=0.475\textwidth]{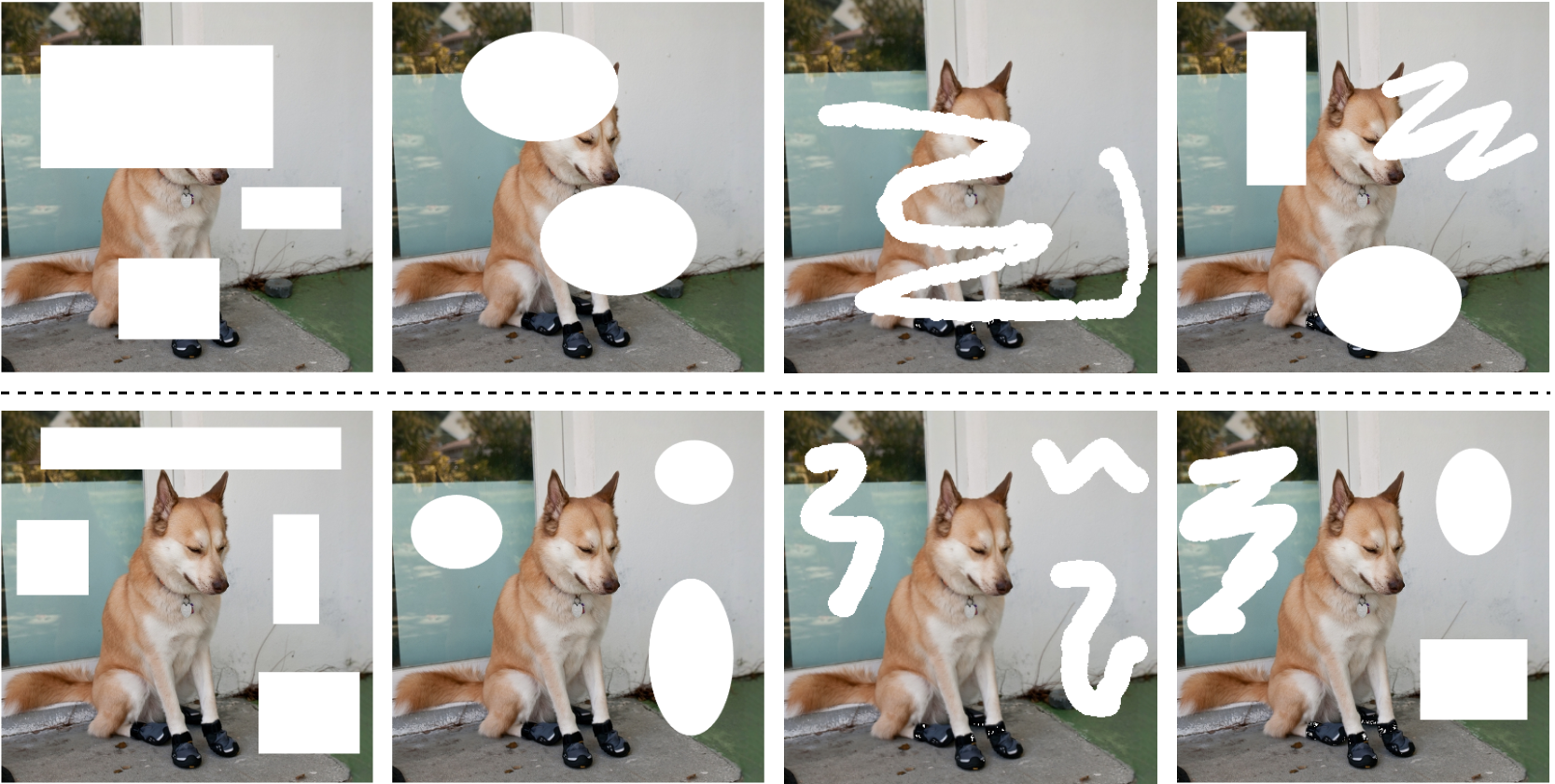}
    \caption{Masking strategies employed in the training processes of StandDiff (first row) and BGDiff (second row).}
    \label{fig:mask}
% \vspace{-0.6em}
\end{figure}

To perform a thorough comparative analysis of the performance between the standard diffusion training method and the erase diffusion training method in the context of the erase inpainting task, we first trained the SD2-Inpaint model (hereafter referred to as StandDiff) using the conventional training approach on the OpenImages V5 dataset~\cite{OpenImages}. The masking strategies employed include rectangular, elliptical, and irregular masks, as well as their random combinations. Additionally, in subsequent experiments, we investigated the impact of constraining the masked regions to the background areas of the original images (designated as BGDiff). This approach is intended to enhance the SD2-Inpaint model's tendency to recover background information during the denoising process, thereby visually improving the effectiveness of eliminating the target object. Figure~\ref{fig:mask} presents the relevant masking strategies, while the other training methodologies remain consistent with those utilized in EraDiff to ensure a fair comparison. 

Table~\ref{tab:suppl:diff} presents the relevant experimental results. The data clearly indicate that the diffusion models trained using these methodologies achieve relatively low and comparable LPIPS scores, suggesting that these models effectively maintain visual coherence in the erased images, which represents a significant advantage of diffusion-based approaches. However, when employing the standard training method, the diffusion model exhibits suboptimal performance in object elimination. Conversely, restricting the masked regions during training to the background areas of the original images leads to a noticeable improvement in the model's object elimination capabilities, although it still falls short compared to EraDiff. 
%%%%%%%
This phenomenon arises from the observation that both StandDiff and BGDiff emphasize the restoration of identical masked regions at each time step during the training process. In contrast, EraDiff introduces a subtle shift in these regions between the time steps \( t \) and \( t-1 \). This variation can be interpreted as a minor perturbation within the masked areas, which enhances the model's capacity for continuous reasoning as opposed to merely reproducing prior outputs. Consequently, during the denoising process, when artifacts manifest in the erasure-target region, EraDiff dynamically adjusts to progressively eliminate these artifacts. Conversely, models trained using previous methodologies tend to rely heavily on and reinforce their descriptions, resulting in high tolerance for error pixels within the model. It is noteworthy that when \( t \) is significantly large, the associated noise will also be substantial, leading to a higher probability of generating artifacts or other non-target content during denoising. Therefore, these high-tolerance models exhibit limited capability in eliminating unwanted objects. The experimental results presented in Figure 3 of the main text further substantiate this observation.

%%%%%
\begin{table}[t]
    \centering
    \begin{tabular}{p{2.7cm} p{1.0cm} p{1.45cm} p{1.5cm}}
        \toprule
        Method & LPIPS & Local FID & GPT score \\
        \midrule
        StandDiff                                   & 0.285                     & 6.981                 & 35.65\%             \\
        BGDiff                                      & 0.259                     & 6.588                 & 51.06\%                \\
        \textbf{EraDiff $w/o\ \text{SRA}$}          & \underline{0.198}         & \underline{4.950}     & \underline{78.54\%}    \\
        \textbf{EraDiff (ours)}                     & \textbf{0.192}            & \textbf{3.799}        & \textbf{83.43\%}    \\
        \bottomrule
    \end{tabular}
    \raggedright % 左对齐
    % \footnotesize{ 
    % $^{\dagger}$ In this experiment, we transformed the class labels corresponding to the masked regions into prompts that serve as guiding information for the denoising process of SD2-Inpaint.
    % }
\caption{Quantitative assessment of different training methods for diffusion models on the OpenImages V5 test set. Optimal results are highlighted in bold, with runner-up performance underlined.}
\label{tab:suppl:diff}
% \vspace{-0.6em}
\end{table}
%%%%%%%
% %%%%%%
% %%%%%%

\section{Supplementary Experiments}
\label{suppl:experi}
% 
%%%%%%
\begin{figure*}[t]
    \centering
    \includegraphics[width=\textwidth]{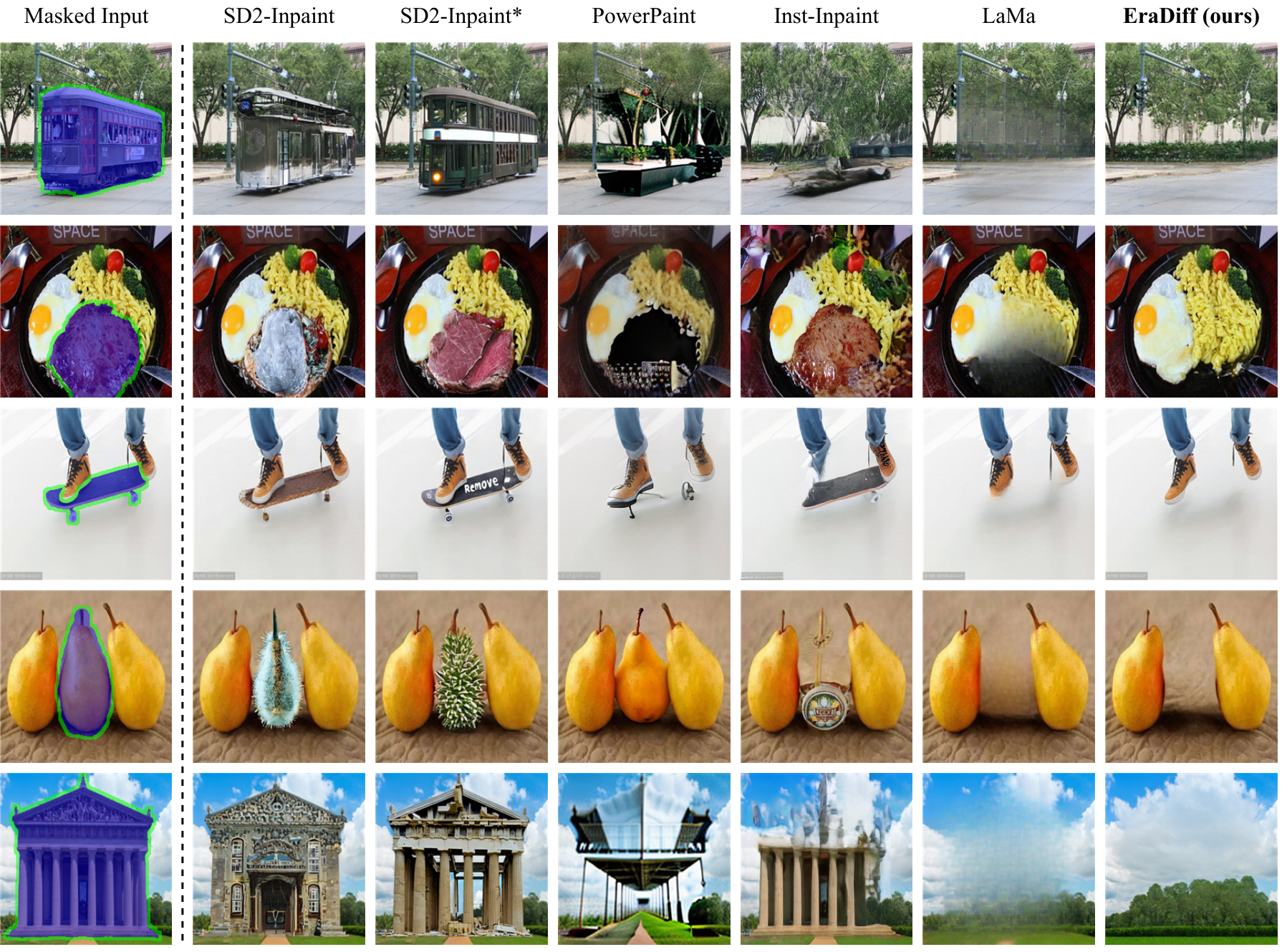}
    \caption{Qualitative results of FSS-1000 dataset compared among SD2-Inpaint~\cite{LDMs}, SD2-Inpaint with prompt guidance~\cite{LDMs}, PowerPaint~\cite{powerpaint}, Inst-Inpaint~\cite{InstInpaint}, LaMa~\cite{LaMa}, and our approach.}
    \label{fig:suppl:vis_coherence}
% \vspace{-0.6em}
\end{figure*}
%%%%%
To enhance the credibility of the comparison results between baseline models and EraDiff, we introduce a new testing dataset, FSS-1000~\cite{fss}. The FSS-1000 dataset encompasses 1,000 distinct categories with a total of 10,000 samples. It features a rich variety of images of animals and everyday objects, in addition to items such as merchandise and logos, which are relatively underrepresented in other existing segmentation datasets.
Regarding evaluation metrics, we utilize LPIPS, FID, and Local FID to measure visual coherence in erased images, which is consistent with the metrics used in the OpenImages V5 dataset~\cite{OpenImages}.In additional experiments, we employ GPT-4o to evaluate the effectiveness of the top three performing models in visual coherence for eliminating erasure targets. All models are compared equitably, without any fine-tuning on the FSS-1000 dataset. Tables~\ref{tab:suppl:main} and~\ref{tab:suppl:gsb} present the relevant experimental results. Additionally, a visual comparison of these methods is illustrated on the FSS-1000 dataset in Figure~\ref{fig:suppl:vis_coherence}. 
%%%%%
\begin{table}[t]
    \centering
    \begin{tabular}{p{2.3cm} p{1.35cm} p{1.3cm} p{1.7cm}}
        \toprule
        Method & FID$ \downarrow $ & LPIPS$ \downarrow $ & Local FID$ \downarrow $\\
        \midrule
        SD2-Inpaint                     & \underline{6.982}     & 0.248                 & 1.201             \\
        SD2-Inpaint$^{*}$               & \textbf{6.874}        & 0.231                 & \underline{1.053} \\
        PowerPaint                      & 12.885                 & 0.395                 & 1.759            \\
        Inst-Inpaint                    & 7.320                & 0.336                 & 2.284            \\
        LaMa                            & 8.093                 & \underline{0.142}     & 1.185                 \\
        \textbf{EraDiff (ours)}         & 7.751                 & \textbf{0.127}       & \textbf{0.869}        \\
        \bottomrule
    \end{tabular}
    \raggedright % 左对齐
    % \footnotesize{ 
    % $^{\dagger}$ In this experiment, we transformed the class labels corresponding to the masked regions into prompts that serve as guiding information for the denoising process of SD2-Inpaint.
    % }
\caption{Quantitative evaluation of baseline models and EraDiff on the FSS-1000 dataset. The optimal results are indicated in bold, and the sub-optimal results are indicated with underlines.}
\label{tab:suppl:main}
% \vspace{-0.7em}
\end{table}
%%%%%%%
%%%%
\begin{table}[t]
% \vspace{-0.2em}
    \centering
    \begin{tabular}{p{2.2cm} p{1.35cm} p{1.7cm} p{1.4cm}}
        \toprule
        % Method & $\mathcal{S}\uparrow$ & $\mathcal{E}$ & $\mathcal{I}\downarrow$\\
        Method & Superior & Comparable & Inferior\\
        \midrule
        SD2-Inpaint              & 2.42\%       & 19.36\%             & 78.22\%         \\
        SD2-Inpaint$^{*}$        & 3.78\%       & 25.97\%             & 70.25\%         \\
        LaMa                     & 12.93\%      & 33.16\%             & 53.91\%         \\
        \bottomrule
    \end{tabular}
    % \raggedright % 左对齐
    % \footnotesize{ 
    % $^{\dagger}$ In this experiment, we transformed the class labels corresponding to the masked regions into prompts that serve as guiding information for the denoising process of SD2-Inpaint.
    % }
\caption{Quantitative results of FSS-1000 dataset among SD2-Inpaint, SD2-Inpaint$^*$, LaMa, and EraDiff. This table delineates a comparative analysis of the elimination performance results obtained by these methodologies relative to ours, highlighting whether their outcomes are superior, comparable, or inferior to those achieved by our approach.}
\label{tab:suppl:gsb}
% \vspace{-0.6em}
\end{table}
%%%%%%%%%%

% As shown in Table~\ref{tab:suppl:main}, the models SD2-Inpaint, SD2-Inpaint*, LaMa, and EraDiff achieve commendable performance in both LPIPS and FID metrics. 
The results presented in Table~\ref{tab:suppl:main} illustrate the performance of the models SD2-Inpaint, SD2-Inpaint*, LaMa, and EraDiff, all of which demonstrate commendable outcomes as assessed through the LPIPS and FID metrics.
Notably, EraDiff attains the best scores in both LPIPS and Local FID, further validating its capability to ensure visual coherence in erased images while maintaining high visual fidelity in the erased regions.
Table \ref{tab:suppl:gsb} offers a comparative analysis of the performance of SD2-Inpaint, SD2-Inpaint*, and LaMa relative to our proposed model, EraDiff, in eliminating specified objects. The results unequivocally demonstrate that these alternative models exhibit significantly inferior performance relative to EraDiff. 
% This disparity can be attributed primarily to the artifacts present in the erased regions of images processed with these techniques, as evidenced by the validation presented in Figure~\ref{fig:suppl:vis_coherence}. 
One of the primary reasons for this performance gap is the presence of undesirable artifacts in the erased regions of images processed by these alternative methods. Such limitations are visually substantiated by the evidence presented in Figure~\ref{fig:suppl:vis_coherence}.
These findings further substantiate that EraDiff, through the calibration of sampling pathways, effectively accomplishes target removal during the task of erase inpainting and excels in achieving both high visual quality and consistency in the generated images.

\section{Generalization in Varied Scenarios}
\label{suppl:gene}
To rigorously evaluate the generalization capability of EraDiff, we conducted comprehensive experiments utilizing two distinct datasets, each exhibiting unique distribution characteristics. The first dataset comprises marketable product images sourced from online e-commerce platforms, while the second dataset includes cartoon character images drawn from the publicly available ATD-12k dataset~\cite{atd}. We meticulously analyzed the performance of EraDiff in comparison to baseline models, with the resulting visualizations provided in Figures~\ref{fig:suppl:suppl_scene1} and~\ref{fig:suppl:suppl_scene2}, respectively. 
These comparisons highlight the efficacy of our proposed model across diverse image distributions.
% thereby demonstrating its robustness and adaptability.

% \section{Qualitative Results of Irregular Masks}
% \label{suppl:mask}
% % 
% Having the supplementar

\section{Experimental Details}
\label{suppl:details}

\begin{figure}[t]
% \vspace{2em}
    \centering
    \includegraphics[width=0.475\textwidth]{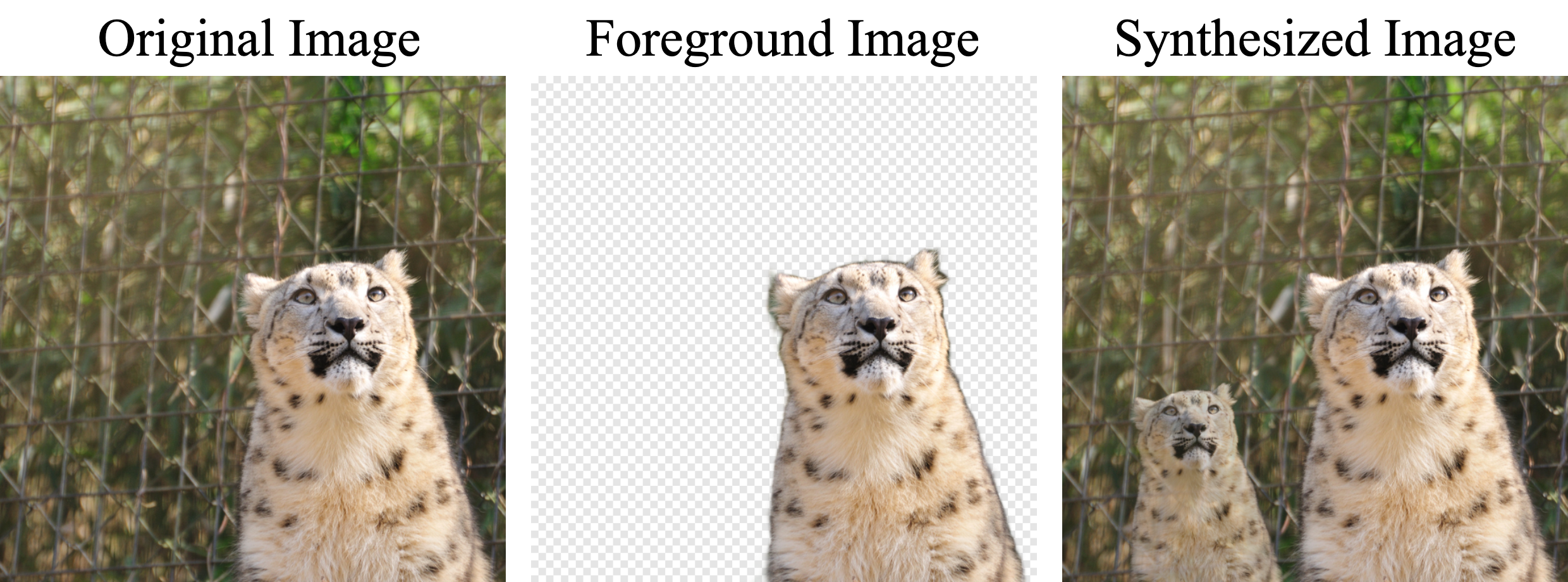}
    \caption{Data synthesis process for model training in this study.}
    \label{fig:suppl:suppl_synthesis}
% \vspace{-0.2em}
\end{figure}

\noindent \textbf{Data synthesis.} As shown in Figure~\ref{fig:suppl:suppl_synthesis}, the data synthesis process required during the training phase is as follows: First, we employed matting techniques~\cite{mask2former,vitmatte} to extract the foreground \(\bm{x}_0^{obj}\) from the \(\bm{x}_0^{ori}\) and obtained a corresponding mask necessary for distinguishing between the foreground and background. Next, we scaled the \(\bm{x}_0^{obj}\) by a random ratio ranging from 50\% to 120\%, followed by a random rotation from 0 to 360 degrees. Finally, we utilized the earlier obtained mask to seamlessly mix-up the modified \(\bm{x}_0^{obj}\) and \(\bm{x}_0^{ori}\) in the background.  

\noindent \textbf{Training.} The training process utilized a batch size of 32 across a cluster of 16 A100 GPUs, for a total of 5 epochs. The noise scheduler was set to DDIM~\cite{ddim}. In the experiment, we fine-tuned the U-Net~\cite{unet} of SD2-Inpaint, while the parameters of the other modules were frozen.

\noindent \textbf{Inference.} 
During the inference phase, we employed the DPMSolverMultistepScheduler~\cite{dpmsolver}. Importantly, we deliberately omitted the use of classifier-free guidance (CFG) and auxiliary prompts to simplify the reverse process. All images in this paper were generated at a resolution of $512 \times 512$ pixels, utilizing 20 denoising steps with a denoising strength of 0.95.

%%%%%%
%%%%%

%%%%%%
\begin{figure}[t]
\vspace{-0.8em}
    \centering
    \includegraphics[width=0.475\textwidth]{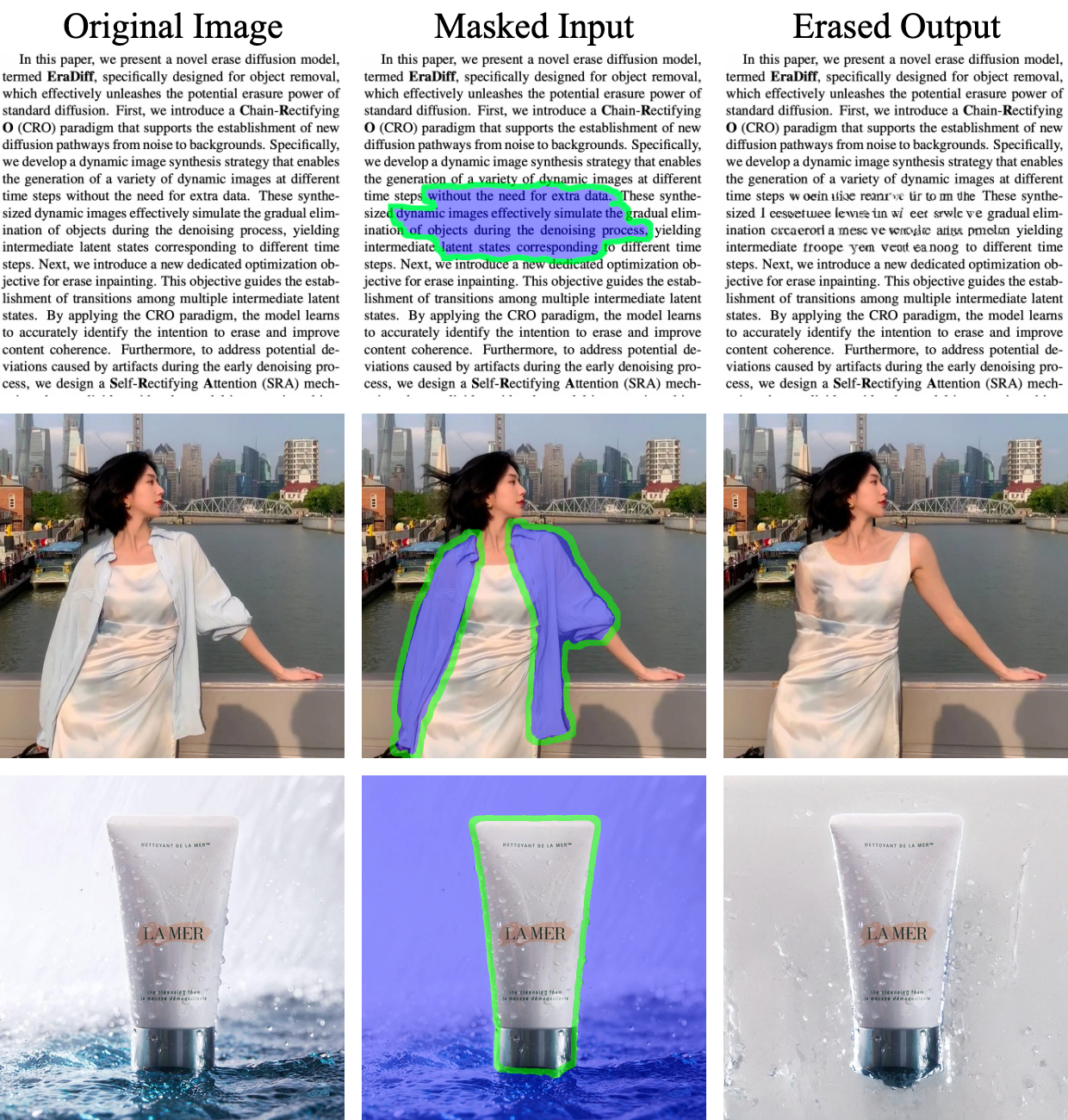}
    \caption{Failure cases of our approach.}
    \label{fig:suppl:suppl_limitation}
\vspace{-0.2em}
\end{figure}
%%%%%

%%%%%%
\begin{figure*}[h]
\vspace{2em}
    \centering
    \includegraphics[width=\textwidth]{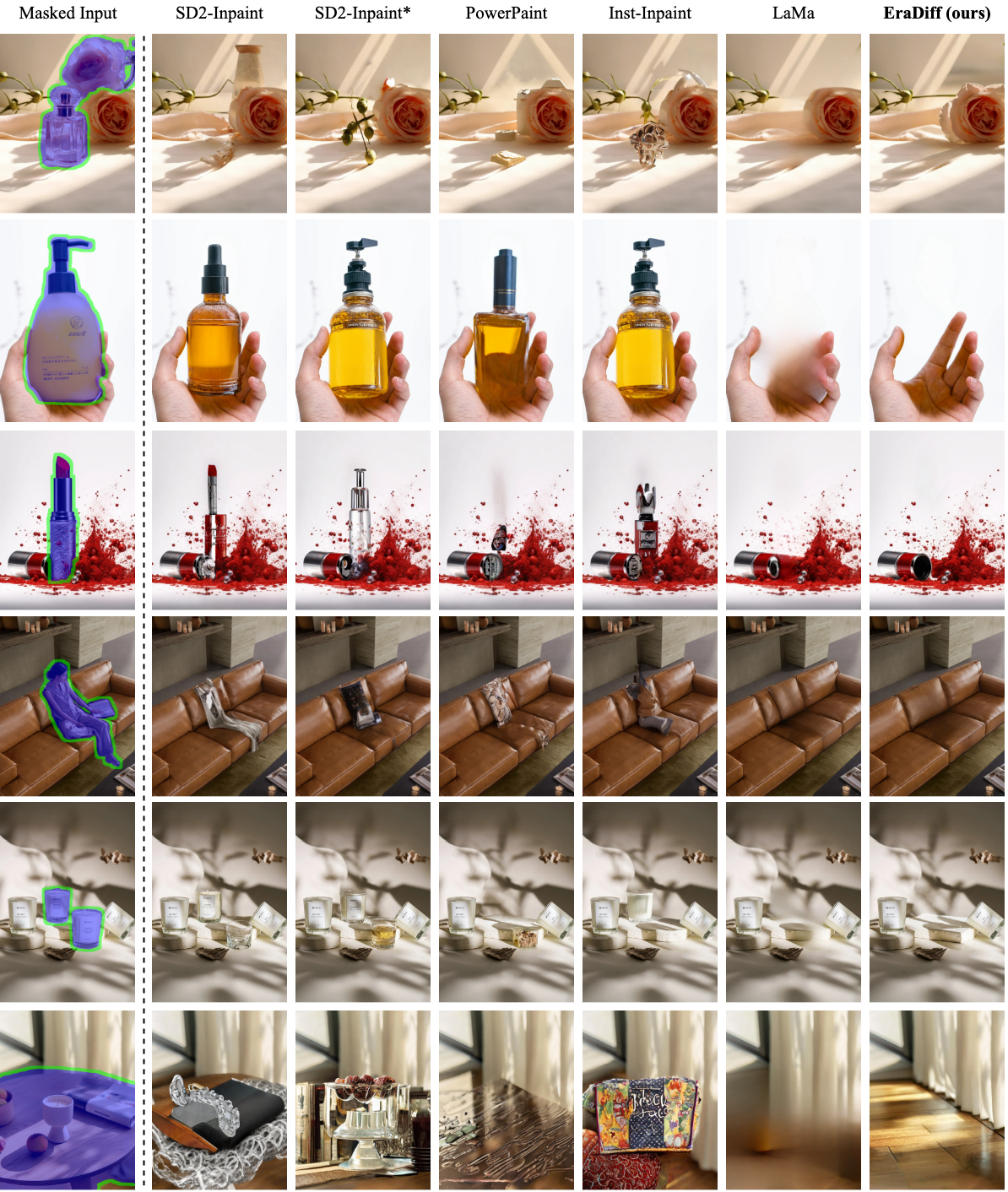}
    \caption{Comparison of visualizations for baseline models and the proposed EraDiff in scenarios of marketable products.}
    \label{fig:suppl:suppl_scene1}
% \vspace{-0.6em}
\end{figure*}
%%%%%

%%%%%%
\begin{figure*}[h]
\vspace{2em}
    \centering
    \includegraphics[width=\textwidth]{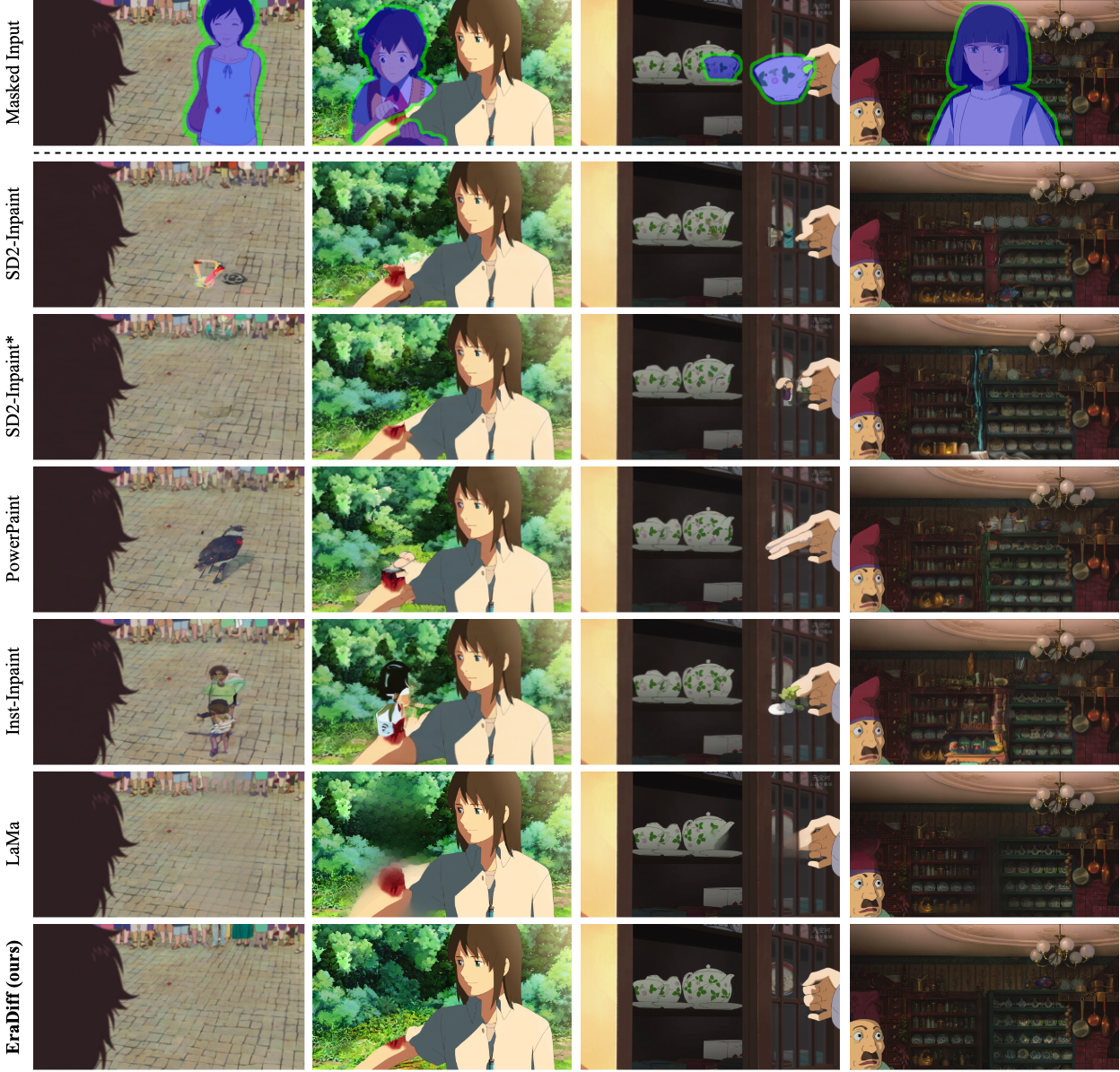}
    \caption{Comparison of visualizations for baseline models and the proposed EraDiff in scenarios of cartoon images.}
    \label{fig:suppl:suppl_scene2}
% \vspace{-0.2em}
\end{figure*}
%%%%%

\section{More Quantitative Results}
\label{suppl:nima}
To comprehensively evaluate EraDiff's performance in object elimination, we created a new test set of 10,000 samples using the original data synthesis method from training. This set is based on the OpenImages V5 dataset, with synthesized images requiring objects to be removed and original images serving as the ground truth for comparison.

In this experiment, we introduced novel evaluation metrics to thoroughly assess the quality of both object elimination and image coherence. Specifically, we employed two aesthetic evaluation metrics, AES~\cite{aes} and NIMA~\cite{nima}, as well as PIDS and UIDS~\cite{CoModGAN}, which measure the distinction between the erased images and the target images.

The experimental results in Table~\ref{tab:suppl:nima} demonstrate that EraDiff outperforms the baseline models. This indicates that EraDiff not only effectively removes objects but also maintains the visual consistency of the resulting images.

%%%%%
\begin{table}[t]
    \centering
    % \small
    % \setlength{\tabcolsep}{4.5pt}
    \begin{tabular}{p{1.9cm} p{1.1cm} p{1.2cm} p{0.9cm} p{1.0cm}}
        \toprule
        Method & PIDS(\%) &UIDS(\%) & AES & NIMA  \\
        \midrule
        SD2-Inpaint             & 13.67     & 23.62     & 4.721     & 5.29     \\
        SD2-Inpaint$^{*}$       & 14.35     & 22.19     & 4.793     & 5.536     \\
        PowerPaint              & 4.36     & 11.28     & 4.814     & 5.205     \\
        Inst-Inpaint            & 0.0     & 0.500     & 4.459     & 5.267     \\
        LaMa                    & 18.04     & 25.92     & 4.616     & 5.605    \\
        \textbf{EraDiff}        & \textbf{25.25}     & \textbf{33.72}     & \textbf{5.082}     & \textbf{5.988}     \\
        \bottomrule
    \end{tabular}
    \raggedright % 左对齐
% \vspace{-1.1em}
\caption{Quantitative evaluation of baseline models and EraDiff.}
\label{tab:suppl:nima}
% \vspace{-2.4em}
\end{table}
%%%%%%%

\begin{table}[t]
    \centering
    \begin{tabular}{p{1.9cm} p{1.2cm} p{1.1cm}}
        \toprule
        Method & Params & RT(s) \\
        \midrule
        SD2-Inpaint             &1.29B     &2.61     \\
        SD2-Inpaint$^{*}$       &1.29B     &2.63     \\
        PowerPaint              &2.08B     &22.86     \\
        Inst-Inpaint            &0.51B     &1.80     \\
        LaMa                    &0.05B     &0.43     \\
        \textbf{EraDiff}        &1.29B     &1.74     \\
        \bottomrule
    \end{tabular}
    \caption{Comparison of model parameters and inference time.}
    \label{tab:suppl:cost}
\end{table}

\section{Complexity Evaluation}
\label{suppl:cost}
In this section, we evaluated the computational complexity of the EraDiff model with that of baseline models. The results of this analysis are presented in Table~\ref{tab:suppl:cost}. We found that EraDiff exhibits a moderate level of both parameter count and inference time. Specifically, because EraDiff shares the same architectural structure as SD2-Inpaint, their parameter counts are identical. However, EraDiff benefits from reduced inference time due to the absence of CFG utilization, enhancing its real-time performance efficiency.

\section{Limitation and Failure Cases}
\label{suppl:limitation}
The EraDiff method encounters certain challenges in some situations, as illustrated in Figure~\ref{fig:suppl:suppl_limitation}. Specifically, for document-type data erasure, it tends to produce text-like artifacts in the target erasure regions due to surrounding texts. Additionally, the method may underperform in tasks involving completion. For instance, removing a coat from an individual and reconstructing the arm beneath may yield suboptimal results. Furthermore, when dealing with large-scale background erasure (background replacement), the absence of reference information often leads to less desirable outcomes. We aim to address and overcome these limitations in future research endeavors.

% \section{Qualitative Results of Irregular Masks}
% \label{suppl:mask}
% % 
% Having the supplementar

\clearpage

%% file: main.bbl
\begin{thebibliography}{60}
\providecommand{\natexlab}[1]{#1}
\providecommand{\url}[1]{\texttt{#1}}
\expandafter\ifx\csname urlstyle\endcsname\relax
  \providecommand{\doi}[1]{doi: #1}\else
  \providecommand{\doi}{doi: \begingroup \urlstyle{rm}\Url}\fi

\bibitem[Altinel et~al.(2018)Altinel, Ozay, and Okatani]{dl_sematic_1}
Fazil Altinel, Mete Ozay, and Takayuki Okatani.
\newblock Deep structured energy-based image inpainting.
\newblock In \emph{ICPR}, pages 423--428. {IEEE} Computer Society, 2018.

\bibitem[Avrahami et~al.(2023)Avrahami, Fried, and Lischinski]{bld}
Omri Avrahami, Ohad Fried, and Dani Lischinski.
\newblock Blended latent diffusion.
\newblock \emph{ACM TOG}, 42\penalty0 (4):\penalty0 149:1--149:11, 2023.

\bibitem[Bar{-}Tal et~al.(2022)Bar{-}Tal, Ofri{-}Amar, Fridman, Kasten, and Dekel]{appl3}
Omer Bar{-}Tal, Dolev Ofri{-}Amar, Rafail Fridman, Yoni Kasten, and Tali Dekel.
\newblock Text2live:text-driven layered image and video editing.
\newblock In \emph{ECCV}, pages 707--723. Springer, 2022.

\bibitem[Barnes et~al.(2009)Barnes, Shechtman, Finkelstein, and Goldman]{PatchMatch}
Connelly Barnes, Eli Shechtman, Adam Finkelstein, and Dan~B. Goldman.
\newblock Patchmatch: a randomized correspondence algorithm for structural image editing.
\newblock \emph{ACM TOG}, 28\penalty0 (3):\penalty0 24, 2009.

\bibitem[Cao et~al.(2023)Cao, Dong, and Fu]{ZITS++}
Chenjie Cao, Qiaole Dong, and Yanwei Fu.
\newblock {ZITS++:} image inpainting by improving the incremental transformer on structural priors.
\newblock \emph{IEEE TPAMI}, 45\penalty0 (10):\penalty0 12667--12684, 2023.

\bibitem[Cheng et~al.(2022)Cheng, Misra, Schwing, Kirillov, and Girdhar]{mask2former}
Bowen Cheng, Ishan Misra, Alexander~G. Schwing, Alexander Kirillov, and Rohit Girdhar.
\newblock Masked-attention mask transformer for universal image segmentation.
\newblock In \emph{CVPR}, 2022.

\bibitem[Ding et~al.(2019)Ding, Ram, and Rodr{\'{\i}}guez]{dl_sematic_2}
Ding Ding, Sundaresh Ram, and Jeffrey~J. Rodr{\'{\i}}guez.
\newblock Image inpainting using nonlocal texture matching and nonlinear filtering.
\newblock \emph{IEEE TIP}, 28\penalty0 (4):\penalty0 1705--1719, 2019.

\bibitem[Dong et~al.(2022)Dong, Cao, and Fu]{ZITS}
Qiaole Dong, Chenjie Cao, and Yanwei Fu.
\newblock Incremental transformer structure enhanced image inpainting with masking positional encoding.
\newblock In \emph{{CVPR}}, pages 11348--11358, 2022.

\bibitem[Esfandarani and Milanfar(2018)]{nima}
Hossein~Talebi Esfandarani and Peyman Milanfar.
\newblock {NIMA:} neural image assessment.
\newblock \emph{{IEEE} Trans. Image Process.}, 27\penalty0 (8):\penalty0 3998--4011, 2018.

\bibitem[Geng et~al.(2024)Geng, Yang, Hang, Li, Gu, Zhang, Bao, Zhang, Li, Hu, Chen, and Guo]{inp2}
Zigang Geng, Binxin Yang, Tiankai Hang, Chen Li, Shuyang Gu, Ting Zhang, Jianmin Bao, Zheng Zhang, Houqiang Li, Han Hu, Dong Chen, and Baining Guo.
\newblock Instructdiffusion: {A} generalist modeling interface for vision tasks.
\newblock In \emph{CVPR}, pages 12709--12720. {IEEE}, 2024.

\bibitem[Hays and Efros(2007)]{HaysE07}
James Hays and Alexei~A. Efros.
\newblock Scene completion using millions of photographs.
\newblock \emph{ACM TOG}, 26\penalty0 (3):\penalty0 4, 2007.

\bibitem[Heusel et~al.(2017)Heusel, Ramsauer, Unterthiner, Nessler, and Hochreiter]{fid}
Martin Heusel, Hubert Ramsauer, Thomas Unterthiner, Bernhard Nessler, and Sepp Hochreiter.
\newblock Gans trained by a two time-scale update rule converge to a local nash equilibrium.
\newblock In \emph{NeurIPS}, pages 6626--6637, 2017.

\bibitem[Huang et~al.(2024)Huang, Xie, Wang, Yuan, Cun, Ge, Zhou, Dong, Huang, Zhang, and Shan]{SmartEdit}
Yuzhou Huang, Liangbin Xie, Xintao Wang, Ziyang Yuan, Xiaodong Cun, Yixiao Ge, Jiantao Zhou, Chao Dong, Rui Huang, Ruimao Zhang, and Ying Shan.
\newblock Smartedit: Exploring complex instruction-based image editing with multimodal large language models.
\newblock In \emph{CVPR}, pages 8362--8371. {IEEE}, 2024.

\bibitem[Iizuka et~al.(2017)Iizuka, Simo{-}Serra, and Ishikawa]{dl_sematic_3}
Satoshi Iizuka, Edgar Simo{-}Serra, and Hiroshi Ishikawa.
\newblock Globally and locally consistent image completion.
\newblock \emph{ACM TOG}, 36\penalty0 (4):\penalty0 107:1--107:14, 2017.

\bibitem[Jain et~al.(2023)Jain, Zhou, Yu, and Shi]{FCF}
Jitesh Jain, Yuqian Zhou, Ning Yu, and Humphrey Shi.
\newblock Keys to better image inpainting: Structure and texture go hand in hand.
\newblock In \emph{{WACV}}, pages 208--217, 2023.

\bibitem[Jampani et~al.(2021)Jampani, Chang, Sargent, Kar, Tucker, Krainin, Kaeser, Freeman, Salesin, Curless, and Liu]{3D}
Varun Jampani, Huiwen Chang, Kyle Sargent, Abhishek Kar, Richard Tucker, Michael Krainin, Dominik Kaeser, William~T. Freeman, David Salesin, Brian Curless, and Ce Liu.
\newblock {SLIDE:} single image 3d photography with soft layering and depth-aware inpainting.
\newblock In \emph{ICCV}, pages 12498--12507. {IEEE}, 2021.

\bibitem[Kingma and Ba(2015)]{Adam}
Diederik~P. Kingma and Jimmy Ba.
\newblock Adam: {A} method for stochastic optimization.
\newblock In \emph{ICLR}, 2015.

\bibitem[Kingma and Welling(2014)]{vae}
Diederik~P. Kingma and Max Welling.
\newblock Auto-encoding variational bayes.
\newblock In \emph{ICLR}, 2014.

\bibitem[K{\"{o}}hler et~al.(2014)K{\"{o}}hler, Schuler, Sch{\"{o}}lkopf, and Harmeling]{dl_1}
Rolf K{\"{o}}hler, Christian~J. Schuler, Bernhard Sch{\"{o}}lkopf, and Stefan Harmeling.
\newblock Mask-specific inpainting with deep neural networks.
\newblock In \emph{German Conference on Pattern Recognition}, pages 523--534. Springer, 2014.

\bibitem[Kuznetsova et~al.(2018)Kuznetsova, Rom, Alldrin, Uijlings, Krasin, Pont{-}Tuset, Kamali, Popov, Malloci, Duerig, and Ferrari]{OpenImages}
Alina Kuznetsova, Hassan Rom, Neil Alldrin, Jasper R.~R. Uijlings, Ivan Krasin, Jordi Pont{-}Tuset, Shahab Kamali, Stefan Popov, Matteo Malloci, Tom Duerig, and Vittorio Ferrari.
\newblock The open images dataset {V4:} unified image classification, object detection, and visual relationship detection at scale.
\newblock \emph{CoRR}, abs/1811.00982, 2018.

\bibitem[Levin et~al.(2003)Levin, Zomet, and Weiss]{LevinZW03}
Anat Levin, Assaf Zomet, and Yair Weiss.
\newblock Learning how to inpaint from global image statistics.
\newblock In \emph{{ICCV}}, pages 305--312, 2003.

\bibitem[Li et~al.(2020{\natexlab{a}})Li, Wang, Zhang, Du, and Tao]{dl_engineer_2}
Jingyuan Li, Ning Wang, Lefei Zhang, Bo Du, and Dacheng Tao.
\newblock Recurrent feature reasoning for image inpainting.
\newblock In \emph{CVPR}, pages 7757--7765. Computer Vision Foundation / {IEEE}, 2020{\natexlab{a}}.

\bibitem[Li et~al.(2024)Li, Zeng, Feng, Gao, Liu, Liu, Li, Tang, Hu, Liu, and Zhang]{ZONE}
Shanglin Li, Bohan Zeng, Yutang Feng, Sicheng Gao, Xiuhui Liu, Jiaming Liu, Lin Li, Xu Tang, Yao Hu, Jianzhuang Liu, and Baochang Zhang.
\newblock {ZONE:} zero-shot instruction-guided local editing.
\newblock In \emph{CVPR}, pages 6254--6263. {IEEE}, 2024.

\bibitem[Li et~al.(2022)Li, Lin, Zhou, Qi, Wang, and Jia]{MAT}
Wenbo Li, Zhe Lin, Kun Zhou, Lu Qi, Yi Wang, and Jiaya Jia.
\newblock {MAT:} mask-aware transformer for large hole image inpainting.
\newblock In \emph{{CVPR}}, pages 10748--10758, 2022.

\bibitem[Li et~al.(2020{\natexlab{b}})Li, Wei, Chen, Tai, and Tang]{fss}
Xiang Li, Tianhan Wei, Yau~Pun Chen, Yu{-}Wing Tai, and Chi{-}Keung Tang.
\newblock {FSS-1000:} {A} 1000-class dataset for few-shot segmentation.
\newblock In \emph{CVPR}, pages 2866--2875. Computer Vision Foundation / {IEEE}, 2020{\natexlab{b}}.

\bibitem[Lu et~al.(2022)Lu, Zhou, Bao, Chen, Li, and Zhu]{dpmsolver}
Cheng Lu, Yuhao Zhou, Fan Bao, Jianfei Chen, Chongxuan Li, and Jun Zhu.
\newblock Dpm-solver: A fast ode solver for diffusion probabilistic model sampling in around 10 steps.
\newblock \emph{arXiv preprint arXiv:2206.00927}, 2022.

\bibitem[Lugmayr et~al.(2022)Lugmayr, Danelljan, Romero, Yu, Timofte, and Gool]{RePaint}
Andreas Lugmayr, Martin Danelljan, Andr{\'{e}}s Romero, Fisher Yu, Radu Timofte, and Luc~Van Gool.
\newblock Repaint: Inpainting using denoising diffusion probabilistic models.
\newblock In \emph{CVPR}, pages 11451--11461. {IEEE}, 2022.

\bibitem[OpenAI(2024)]{gpt}
OpenAI.
\newblock Gpt-4o.
\newblock \url{https://openai.com/index/hello-gpt-4o/}, 2024.
\newblock Accessed: 2024-07-10.

\bibitem[Pathak et~al.(2016)Pathak, Kr{\"{a}}henb{\"{u}}hl, Donahue, Darrell, and Efros]{dl_4}
Deepak Pathak, Philipp Kr{\"{a}}henb{\"{u}}hl, Jeff Donahue, Trevor Darrell, and Alexei~A. Efros.
\newblock Context encoders: Feature learning by inpainting.
\newblock In \emph{CVPR}, pages 2536--2544. {IEEE} Computer Society, 2016.

\bibitem[Ramesh et~al.(2022)Ramesh, Dhariwal, Nichol, Chu, and Chen]{mo_ldm2}
Aditya Ramesh, Prafulla Dhariwal, Alex Nichol, Casey Chu, and Mark Chen.
\newblock Hierarchical text-conditional image generation with {CLIP} latents.
\newblock \emph{CoRR}, abs/2204.06125, 2022.

\bibitem[Ren et~al.(2015)Ren, Xu, Yan, and Sun]{dl_2}
Jimmy S.~J. Ren, Li Xu, Qiong Yan, and Wenxiu Sun.
\newblock Shepard convolutional neural networks.
\newblock In \emph{NeurIPS}, pages 901--909, 2015.

\bibitem[Rombach et~al.(2022)Rombach, Blattmann, Lorenz, Esser, and Ommer]{LDMs}
Robin Rombach, Andreas Blattmann, Dominik Lorenz, Patrick Esser, and Bj{\"{o}}rn Ommer.
\newblock High-resolution image synthesis with latent diffusion models.
\newblock In \emph{CVPR}, pages 10674--10685. {IEEE}, 2022.

\bibitem[Ronneberger et~al.(2015)Ronneberger, Fischer, and Brox]{unet}
Olaf Ronneberger, Philipp Fischer, and Thomas Brox.
\newblock U-net: Convolutional networks for biomedical image segmentation.
\newblock In \emph{MICCAI}, pages 234--241, 2015.

\bibitem[Sargsyan et~al.(2023)Sargsyan, Navasardyan, Xu, and Shi]{MI-GAN}
Andranik Sargsyan, Shant Navasardyan, Xingqian Xu, and Humphrey Shi.
\newblock {MI-GAN:} {A} simple baseline for image inpainting on mobile devices.
\newblock In \emph{{ICCV}}, pages 7301--7311, 2023.

\bibitem[Schuhmann(2023)]{aes}
Christoph Schuhmann.
\newblock Aesthetic predictor.
\newblock \url{https://github.com/christophschuhmann/improved-aesthetic-predictor}, 2023.

\bibitem[Sheynin et~al.(2024)Sheynin, Polyak, Singer, Kirstain, Zohar, Ashual, Parikh, and Taigman]{EmuEdit}
Shelly Sheynin, Adam Polyak, Uriel Singer, Yuval Kirstain, Amit Zohar, Oron Ashual, Devi Parikh, and Yaniv Taigman.
\newblock Emu edit: Precise image editing via recognition and generation tasks.
\newblock In \emph{CVPR}, pages 8871--8879. {IEEE}, 2024.

\bibitem[Siyao et~al.(2021)Siyao, Zhao, Yu, Sun, Metaxas, Loy, and Liu]{atd}
Li Siyao, Shiyu Zhao, Weijiang Yu, Wenxiu Sun, Dimitris Metaxas, Chen~Change Loy, and Ziwei Liu.
\newblock Deep animation video interpolation in the wild.
\newblock In \emph{CVPR}, 2021.

\bibitem[Song et~al.(2021{\natexlab{a}})Song, Meng, and Ermon]{ddim}
Jiaming Song, Chenlin Meng, and Stefano Ermon.
\newblock Denoising diffusion implicit models.
\newblock In \emph{ICLR}. OpenReview.net, 2021{\natexlab{a}}.

\bibitem[Song et~al.(2021{\natexlab{b}})Song, Sohl{-}Dickstein, Kingma, Kumar, Ermon, and Poole]{mo_ldm1}
Yang Song, Jascha Sohl{-}Dickstein, Diederik~P. Kingma, Abhishek Kumar, Stefano Ermon, and Ben Poole.
\newblock Score-based generative modeling through stochastic differential equations.
\newblock In \emph{ICLR}. OpenReview.net, 2021{\natexlab{b}}.

\bibitem[Sun et~al.(2025)Sun, Cui, Dong, and Tang]{aas}
Wenhao Sun, Benlei Cui, Xue-Mei Dong, and Jingqun Tang.
\newblock Attentive eraser: Unleashing diffusion model’s object removal potential via self-attention redirection guidance.
\newblock In \emph{AAAI}, 2025.

\bibitem[Suvorov et~al.(2022)Suvorov, Logacheva, Mashikhin, Remizova, Ashukha, Silvestrov, Kong, Goka, Park, and Lempitsky]{LaMa}
Roman Suvorov, Elizaveta Logacheva, Anton Mashikhin, Anastasia Remizova, Arsenii Ashukha, Aleksei Silvestrov, Naejin Kong, Harshith Goka, Kiwoong Park, and Victor Lempitsky.
\newblock Resolution-robust large mask inpainting with fourier convolutions.
\newblock In \emph{{WACV}}, pages 3172--3182, 2022.

\bibitem[Wang et~al.(2023)Wang, Saharia, Montgomery, Pont{-}Tuset, Noy, Pellegrini, Onoe, Laszlo, Fleet, Soricut, Baldridge, Norouzi, Anderson, and Chan]{mo_ldm3}
Su Wang, Chitwan Saharia, Ceslee Montgomery, Jordi Pont{-}Tuset, Shai Noy, Stefano Pellegrini, Yasumasa Onoe, Sarah Laszlo, David~J. Fleet, Radu Soricut, Jason Baldridge, Mohammad Norouzi, Peter Anderson, and William Chan.
\newblock Imagen editor and editbench: Advancing and evaluating text-guided image inpainting.
\newblock In \emph{CVPR}, pages 18359--18369. {IEEE}, 2023.

\bibitem[Wang et~al.(2024)Wang, Wu, Zhang, and Xu]{appl2}
Yuxin Wang, Qianyi Wu, Guofeng Zhang, and Dan Xu.
\newblock Learning 3d geometry and feature consistent gaussian splatting for object removal.
\newblock In \emph{ECCV}, pages 1--17. Springer, 2024.

\bibitem[Winter et~al.(2024)Winter, Cohen, Fruchter, Pritch, Rav{-}Acha, and Hoshen]{appl1}
Daniel Winter, Matan Cohen, Shlomi Fruchter, Yael Pritch, Alex Rav{-}Acha, and Yedid Hoshen.
\newblock Objectdrop: Bootstrapping counterfactuals for photorealistic object removal and insertion.
\newblock In \emph{ECCV}, pages 112--129. Springer, 2024.

\bibitem[Xie et~al.(2023{\natexlab{a}})Xie, Zhang, Lin, Hinz, and Zhang]{SmartBrush}
Shaoan Xie, Zhifei Zhang, Zhe Lin, Tobias Hinz, and Kun Zhang.
\newblock Smartbrush: Text and shape guided object inpainting with diffusion model.
\newblock In \emph{CVPR}, pages 22428--22437. {IEEE}, 2023{\natexlab{a}}.

\bibitem[Xie et~al.(2023{\natexlab{b}})Xie, Zhang, Lin, Hinz, and Zhang]{local_fid}
Shaoan Xie, Zhifei Zhang, Zhe Lin, Tobias Hinz, and Kun Zhang.
\newblock Smartbrush: Text and shape guided object inpainting with diffusion model.
\newblock In \emph{CVPR}, pages 22428--22437. {IEEE}, 2023{\natexlab{b}}.

\bibitem[Xie et~al.(2023{\natexlab{c}})Xie, Zhao, Xiao, Chan, Li, Xu, Zhang, and Hou]{DreamInpainter}
Shaoan Xie, Yang Zhao, Zhisheng Xiao, Kelvin C.~K. Chan, Yandong Li, Yanwu Xu, Kun Zhang, and Tingbo Hou.
\newblock Dreaminpainter: Text-guided subject-driven image inpainting with diffusion models.
\newblock \emph{CoRR}, abs/2312.03771, 2023{\natexlab{c}}.

\bibitem[Xu et~al.(2023)Xu, Navasardyan, Tadevosyan, Sargsyan, Mu, and Shi]{SH-GAN}
Xingqian Xu, Shant Navasardyan, Vahram Tadevosyan, Andranik Sargsyan, Yadong Mu, and Humphrey Shi.
\newblock Image completion with heterogeneously filtered spectral hints.
\newblock In \emph{{WACV}}, pages 4580--4590, 2023.

\bibitem[Yang et~al.(2023{\natexlab{a}})Yang, Zhang, Ma, Liu, Fu, and He]{MAGIC}
Siyuan Yang, Lu Zhang, Liqian Ma, Yu Liu, Jingjing Fu, and You He.
\newblock Magicremover: Tuning-free text-guided image inpainting with diffusion models.
\newblock \emph{CoRR}, abs/2310.02848, 2023{\natexlab{a}}.

\bibitem[Yang et~al.(2023{\natexlab{b}})Yang, Zhang, Ma, Liu, Fu, and He]{Magicremover}
Siyuan Yang, Lu Zhang, Liqian Ma, Yu Liu, Jingjing Fu, and You He.
\newblock Magicremover: Tuning-free text-guided image inpainting with diffusion models.
\newblock \emph{CoRR}, abs/2310.02848, 2023{\natexlab{b}}.

\bibitem[Yao et~al.(2024)Yao, Wang, Yang, and Wang]{vitmatte}
Jingfeng Yao, Xinggang Wang, Shusheng Yang, and Baoyuan Wang.
\newblock Vitmatte: Boosting image matting with pre-trained plain vision transformers.
\newblock \emph{Information Fusion}, 103:\penalty0 102091, 2024.

\bibitem[Yi et~al.(2020)Yi, Tang, Azizi, Jang, and Xu]{HiFill}
Zili Yi, Qiang Tang, Shekoofeh Azizi, Daesik Jang, and Zhan Xu.
\newblock Contextual residual aggregation for ultra high-resolution image inpainting.
\newblock In \emph{{CVPR}}, pages 7505--7514, 2020.

\bibitem[Yildirim et~al.(2023)Yildirim, Baday, Erdem, Erdem, and Dundar]{InstInpaint}
Ahmet~Burak Yildirim, Vedat Baday, Erkut Erdem, Aykut Erdem, and Aysegul Dundar.
\newblock Inst-inpaint: Instructing to remove objects with diffusion models.
\newblock \emph{CoRR}, abs/2304.03246, 2023.

\bibitem[Yoon and Cho(2024)]{appl6}
Donggeun Yoon and Donghyeon Cho.
\newblock {CORE-MPI:} consistency object removal with embedding multiplane image.
\newblock In \emph{CVPR}, pages 20081--20090. {IEEE}, 2024.

\bibitem[Zhang et~al.(2018)Zhang, Isola, Efros, Shechtman, and Wang]{lpips}
Richard Zhang, Phillip Isola, Alexei~A. Efros, Eli Shechtman, and Oliver Wang.
\newblock The unreasonable effectiveness of deep features as a perceptual metric.
\newblock In \emph{CVPR}, pages 586--595, 2018.

\bibitem[Zhao et~al.(2021)Zhao, Cui, Sheng, Dong, Liang, Chang, and Xu]{CoModGAN}
Shengyu Zhao, Jonathan Cui, Yilun Sheng, Yue Dong, Xiao Liang, Eric~I{-}Chao Chang, and Yan Xu.
\newblock Large scale image completion via co-modulated generative adversarial networks.
\newblock In \emph{ICLR}, 2021.

\bibitem[Zhou et~al.(2019)Zhou, Hu, Zhang, and Shen]{zhou2019visual}
Hao Zhou, Chuanping Hu, Chongyang Zhang, and Shengyang Shen.
\newblock Visual relationship recognition via language and position guided attention.
\newblock In \emph{ICASSP}, 2019.

\bibitem[Zhou et~al.(2021)Zhou, Zhang, Luo, Chen, and Hu]{zhou2021embracing}
Hao Zhou, Chongyang Zhang, Yan Luo, Yanjun Chen, and Chuanping Hu.
\newblock Embracing uncertainty: Decoupling and de-bias for robust temporal grounding.
\newblock In \emph{CVPR}, pages 8445--8454, 2021.

\bibitem[Zhou et~al.(2020)Zhou, Ding, Lin, Wang, and Tao]{dl_engineer_1}
Tong Zhou, Changxing Ding, Shaowen Lin, Xinchao Wang, and Dacheng Tao.
\newblock Learning oracle attention for high-fidelity face completion.
\newblock In \emph{CVPR}, pages 7677--7686, 2020.

\bibitem[Zhuang et~al.(2023)Zhuang, Zeng, Liu, Yuan, and Chen]{powerpaint}
Junhao Zhuang, Yanhong Zeng, Wenran Liu, Chun Yuan, and Kai Chen.
\newblock A task is worth one word: Learning with task prompts for high-quality versatile image inpainting.
\newblock \emph{CoRR}, abs/2312.03594, 2023.

\end{thebibliography}
